\documentclass[utf8]{article} 

\usepackage{arxiv}

\usepackage{url,hyperref,microtype,subcaption}
\usepackage[onehalfspacing]{setspace}

\usepackage{makecell}
\usepackage[nolist]{acronym}
\usepackage{tabularx}
\usepackage{adjustbox}
\usepackage{booktabs}
\usepackage{natbib}

\title{Interactive Imitation Learning for Dexterous Robotic Manipulation: Challenges and Perspectives—A Survey}

\author{
 Edgar Welte \\
  Karlsruhe Institute of Technology (KIT)\\
  Karlsruhe, Germany \\
  \texttt{edgar.welte@kit.edu} \\
  \And
 Rania Rayyes \\
   Karlsruhe Institute of Technology (KIT)\\
  Karlsruhe, Germany \\
}

\begin{document}
\maketitle

\begin{abstract}

Dexterous manipulation is a crucial yet highly complex challenge in humanoid robotics, demanding precise, adaptable, and sample-efficient learning methods. As humanoid robots are usually designed to operate in human-centric environments and interact with everyday objects, mastering dexterous manipulation is critical for real-world deployment. Traditional approaches, such as reinforcement learning and imitation learning, have made significant strides, but they often struggle due to the unique challenges of real-world dexterous manipulation, including high-dimensional control, limited training data, and covariate shift. This survey provides a comprehensive overview of these challenges and reviews existing learning-based methods for real-world dexterous manipulation, spanning imitation learning, reinforcement learning, and hybrid approaches. A promising yet underexplored direction is interactive imitation learning, where human feedback actively refines a robot’s behavior during training. While interactive imitation learning has shown success in various robotic tasks, its application to dexterous manipulation remains limited. To address this gap, we examine current interactive imitation learning techniques applied to other robotic tasks and discuss how these methods can be adapted to enhance dexterous manipulation. By synthesizing state-of-the-art research, this paper highlights key challenges, identifies gaps in current methodologies, and outlines potential directions for leveraging interactive imitation learning to improve dexterous robotic skills.

\end{abstract}

\section{Introduction}
Recent advances in robot hardware and learning algorithms have led to a surge of interest in dexterous manipulation as a key area of robotics research. Whether in the context of humanoid robots interacting in human environments, or robotic hands performing precise object manipulations, dexterous manipulation presents unique challenges due to its high-dimensional action spaces, complex kinematics, and intricate contact dynamics \citep{Zhu2019DexterousLow-Cost, Sampath2023Reviewhands, Yu2022DexterousReview}. These factors make learning-based approaches notably appealing. 
As the dimensionality of the action space increases, the amount of training data required grows exponentially \citep{Sutton2018ReinforcementIntroduction, Lu2024OvercomingFactorization, Kubus2018LearningEfficiently}, making sample-efficient learning methods increasingly crucial.
Imitation learning has emerged as an effective strategy for dexterous manipulation, enabling robots to learn complex skills by mimicking human demonstrations \citep{Mikami2009ImitationLearning}. Leveraging recorded training data to learn a policy via imitation learning offers high sample efficiency, especially compared to reinforcement learning approaches where the policy is developed independently through interaction with the environment \citep{Radosavovic2021State-OnlyManipulation, Hu2023REBOOTManipulation}. This efficiency is particularly valuable in real-world scenarios, where frequent and random interactions with the environment can be both hazardous and costly \citep{Sutton2018ReinforcementIntroduction, Wang2022AnTasks, Han2023AManipulation}.
However, supervised imitation learning like \ac{BC} is known to suffer from a covariate shift, leading to a mismatch between the state distribution in the training data and the distribution encountered during the execution of the trained policy \citep{Sun2023MEGA-DAggerExperts}. 
\ac{IIL} offers a promising solution to address this challenge by integrating real-time human feedback into the learning process, effectively combining imitation learning with interactive machine learning techniques \citep{Celemin2022InteractiveSurvey}.
Unlike standard imitation learning, which passively learns from fixed demonstrations, \ac{IIL} allows human teachers to actively refine policies by correcting mistakes as they occur during execution. In practice, a policy is pre-trained by imitation learning, and then the policy is executed in the real world. During this execution, human teachers can apply corrections in the event of errors, which allows the policy to adjust accordingly
\citep{Kelly2019HG-DAggerExperts, Celemin2022InteractiveSurvey}. In literature, incorporating human interventions in training is also referred to as a human-in-the-loop approach \citep{Celemin2022InteractiveSurvey, Mandlekar2020Human-in-the-LoopTeleoperation, Wang2024DexCapManipulation}. This human-in-the-loop approach ensures adaptability, enhances sample efficiency, and mitigates covariate shift by dynamically guiding the learning process. It is essential to note that, in our context, the interactive component of \ac{IIL} refers explicitly to human feedback, rather than interactions with the environment, which may also be encompassed by broader interpretations of interactivity in learning. Hence, we believe \ac{IIL} is promising for real-world dexterous manipulation applications.

\begin{figure}[ht]
    \centering
    \includegraphics[width=\textwidth]{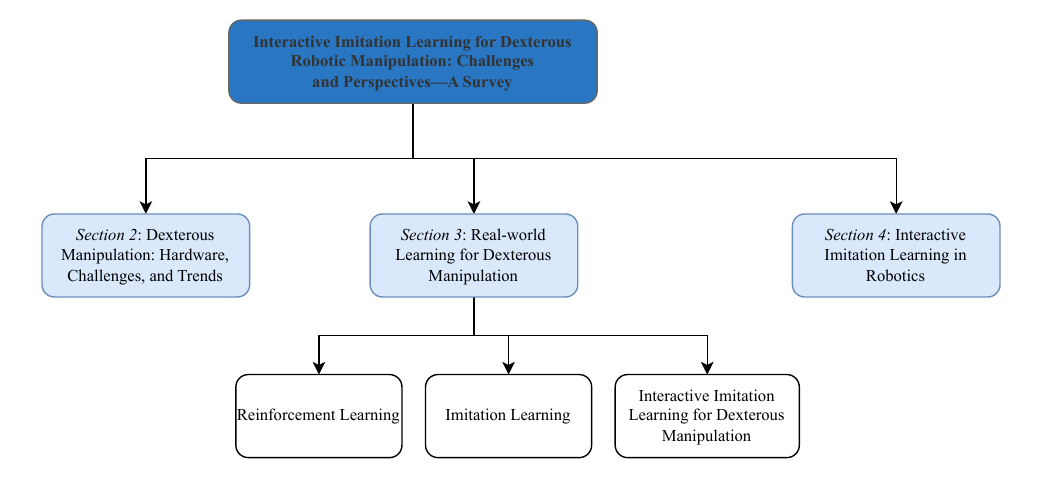}
    \caption{Structural overview of the survey paper on interactive imitation learning for dexterous robotic manipulation.}
    \label{fig:overview}
\end{figure}

This survey provides a comprehensive overview of current approaches in \acl{IIL} for sample efficient real-world dexterous manipulation. To give an intuitive understanding, we broaden our perspective to cover both directions: Dexterous Manipulation and \acl{IIL}. The organization of the paper is illustrated in Fig.\ref{fig:overview}. Section~2 addresses the challenges and trends in dexterous manipulation, laying an essential groundwork for understanding this area of robotics. Section~3 delves into learning-based methods for real-world dexterous manipulation, exploring approaches from imitation learning and reinforcement learning to \ac{IIL}. A broader examination of \ac{IIL}, evaluating its applicability to dexterous manipulation, is discussed in Section~4.

\section{Dexterous Manipulation: Hardware, Challenges, and Trends}

Dexterous Manipulation is a specialized field in robotics focused on controlling multi-fingered end effectors to grasp and manipulate objects effectively \citep{Okamura2000Anmanipulation}. Anthropomorphic robot hands are designed to replicate human activities, offering great versatility. The most comprehensive models of the human hand typically incorporate 20 to 25 \ac{DoF}, with each finger generally modeled with 4 \ac{DoF} and the thumb with 4 or 5 \ac{DoF}. Additionally, the palm and wrist are sometimes included with extra \ac{DoF} to enhance the model's fidelity \citep{Zarzoura2019Investigationfunctions, Savescu2004ASimulation}. Despite their sophistication, anthropomorphic robotic hands face significant technical challenges, particularly in control accuracy, sensor and actuator system dimensioning, and the transmission of power and signals. Alternatively, a minimalist approach can address many of these issues using underactuated hands. These hands have fewer actuators than joints, relying on passive mechanisms like springs or tendons to simplify control and adapt to different object shapes \citep{Birglen2008Underactuatedhands}. Due to the inherent technical and control complexities, the availability of commercially produced anthropomorphic robotic hands was previously limited. Table~\ref{tab:robot_hands} provides an overview of commercially available anthropomorphic robotic hands. Among them, 2/3 are nearly fully actuated, offering a level of dexterity that closely resembles human capabilities. This represents an increase compared to previous years, reflecting the rapidly evolving landscape of dexterous manipulation technologies. In addition to the hands listed, several companies—such as Figure AI\footnote{\url{https://www.figure.ai/}}, 1X\footnote{\url{https://www.1x.tech/}}, and Tesla\footnote{\url{https://www.tesla.com/en_eu/AI}}—are actively developing proprietary robotic hands as part of their humanoid platforms. However, these designs are not yet openly available for research or third-party development.

\begin{table}[ht]
    \caption{Commercial Anthropomorphic Robotic Hands}
    \label{tab:robot_hands}
    \begin{center}
    \adjustbox{max width=\textwidth}{
        \begin{tabular}{lcccccccc}
            \toprule
            \textbf{Name} & \textbf{DoA} & \textbf{DoF} & \textbf{Actuation} & \makecell{\textbf{No. of} \\ \textbf{Fingers}} & 
            \makecell{\textbf{Fingertip} \\ \textbf{Force}} & 
            \textbf{Payload} & \textbf{Weight} & \makecell{\textbf{Tactile} \\ \textbf{Sensors}} \\
            \midrule
            
            \makecell[l]{Shadow Dexterous Hand \\ \small\citep{ShadowHand}} & 20 & 24 & tendon-driven & 5 & / & 5 kg & 4.3 kg & yes \\ [0.4cm]

            \makecell[l]{TESOLLO DG-5F \\ \small\citep{tesolloDG5FHumanoidRobotic2025}} & 20 & 20 & direct-drive & 5 & / & 2.5 - 10 kg & 1.7 kg & yes \\ [0.4cm]
            
            \makecell[l]{Agile Hand \\ \small\citep{AgileHand}} & 16 & 20 & mechanical & 5 & 10 N & / & 1.5 kg & no \\ [0.4cm]
    
            \makecell[l]{ARTUS Lite \\ \small\citep{Artus}} & 16 & 20 & tendon-driven & 5 & 1.5 kg & 5 kg & 1.1 kg & no \\ [0.4cm]
    
            \makecell[l]{mimic hand \\ \small\citep{mimic} \\ \small\citep{Toshimitsu2023GettingJoints}} & 16 & 20 & tendon-driven & 5 & / & 7 kg & 1.1 kg & no \\ [0.6cm]

            \makecell[l]{Unitree Dex5-1 \\ \small\citep{unitreeUnitreeDex51Smart2025}} & 16 & 20 & mechanical & 5 & 10 N & 3.5 - 4.5 kg & 1.0 kg & yes \\ [0.4cm]
            
            \makecell[l]{Allegro Hand V4 / V5 \\ \small\citep{AllegroHand}} & 16 & 16 & direct drive & 4 & / & 5 kg / 15 kg & 1.0 kg & yes \\ [0.4cm]
    
            \makecell[l]{LEAP Hand \\ \small\citep{Shaw2023LEAPLearning}} & 16 & 16 & direct drive & 4 & / & / & / & no \\ [0.4cm]
    
            \makecell[l]{PaXini DexH13GEN2 \\ \small\citep{PaXiniDexH13GEN2}} & 13 & 16 & direct drive & 4 & 15 N & 5 kg & / & yes \\ [0.4cm]
            
            \makecell[l]{XHAND1 \\ \small\citep{XHAND1}} & 12 & 12 & direct drive & 5 & 15 N & 16--25 kg & 1.1 kg & yes \\ [0.4cm]
            
            \makecell[l]{Schunk SVH \\ \small\citep{SchunkSVH}} & 9 & 20 & mechanical & 5 & / & 0.85 kg & 1.3 kg & no \\ [0.4cm]
            
            \makecell[l]{RH8D Robot Hand \\ \small\citep{SeedHand}} & 8 & 19 & tendon-driven & 5 & / & 1--2.5 kg & 620 g & yes \\ [0.4cm]
    
            \makecell[l]{RH56BFX / RH56DFX \\ \small\citep{InspireHand}} & 6 & 12 & mechanical & 5 & 4 N / 10 N & / & 540 g & no \\ [0.4cm]
    
            \makecell[l]{IH2 Azzurra Hand \\ \small\citep{PrensiliaHand}} & 5 & 11 & tendon-driven & 5 & 7 N & / & 640 g & no \\ [0.4cm]
    
            \makecell[l]{qb SoftHand2 Research \\ \small\citep{SoftHand2}} & 2 & 19 & tendon-driven & 5 & / & 2--3 kg & 940 g & no \\ [0.2cm]
            
            \bottomrule
            
        \end{tabular}
    }
    
    \footnotesize Degrees of Actuation (DoA), Degrees of Freedom (DoF), Payload depends on measurement method, tactile sensors include configurable options
    \end{center}
\end{table}

Robotic hands use different actuation methods to control movement, primarily mechanical links, tendon-driven, and direct drive. Mechanical links use rigid components like gears and levers to transmit force, offering precision but at the cost of bulkiness and reduced flexibility. Tendon-driven systems resemble human anatomy by using flexible tendons or cables to control joints, providing lightweight, adaptive, and fluid movements, though they require careful maintenance and calibration. Direct drive places actuators directly at each joint, ensuring highly accurate control with minimal mechanical play, but can be heavy and power-intensive, making it less suitable for compact designs. Each method has trade-offs between precision, adaptability, and complexity \citep{Melchiorri2016RobotHands}. 

While an underactuated robotic hand is generally sufficient for basic tasks, such as picking up and placing household objects \citep{Gross2024OPENGRASP-LITEMechanism}, more intricate operations, such as in-hand manipulation or handling small objects, demand a robotic hand with enhanced agility. For example, an impressive demonstration of in-hand manipulation was presented by \citet{Akkaya2019SolvingHand}, where one Shadow Dexterous Hand solved a Rubik's Cube entirely within its grasp.

\begin{figure}[ht]
    \centering
    \includegraphics[width=\textwidth]{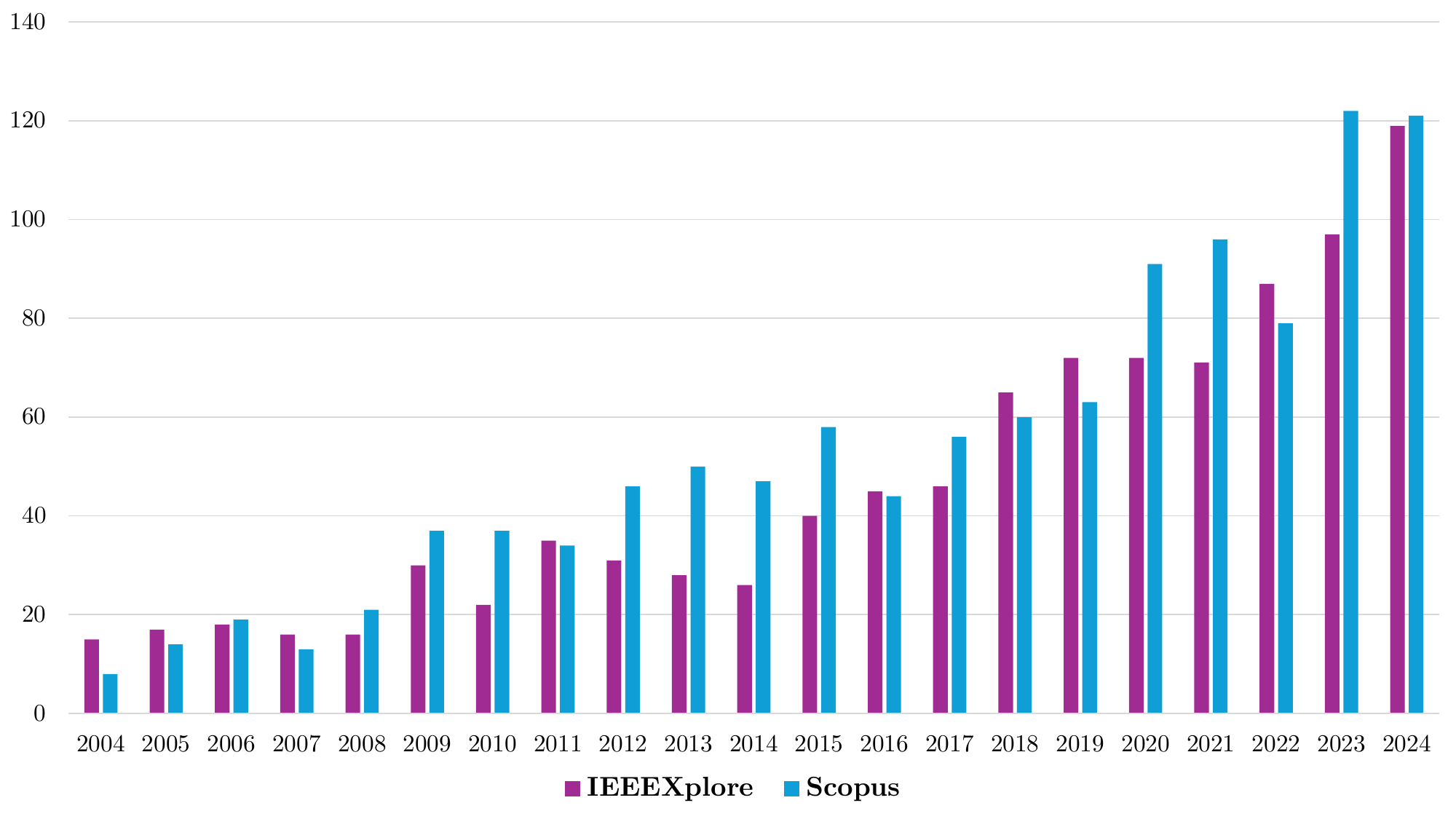}
    \caption{Number of publications with the keyword "dexterous manipulation" over the past 20 years on the databases Scopus and IEEEXplore (checked on 05.08.2025).}
    \label{fig:publications_dex_man}
\end{figure}

With the availability of hardware and the progress of learning control algorithms, research in dexterous manipulation has become increasingly popular in recent years. Figure \ref{fig:publications_dex_man} shows the number of publications in the databases Scopus\footnote{\url{https://www.scopus.com/}} and IEEEXplore\footnote{\url{https://ieeexplore.ieee.org/}} tagged with the keyword “dexterous manipulation” over the past 20 years.

\begin{figure}[ht]
    \centering
    \includegraphics[width=\textwidth]{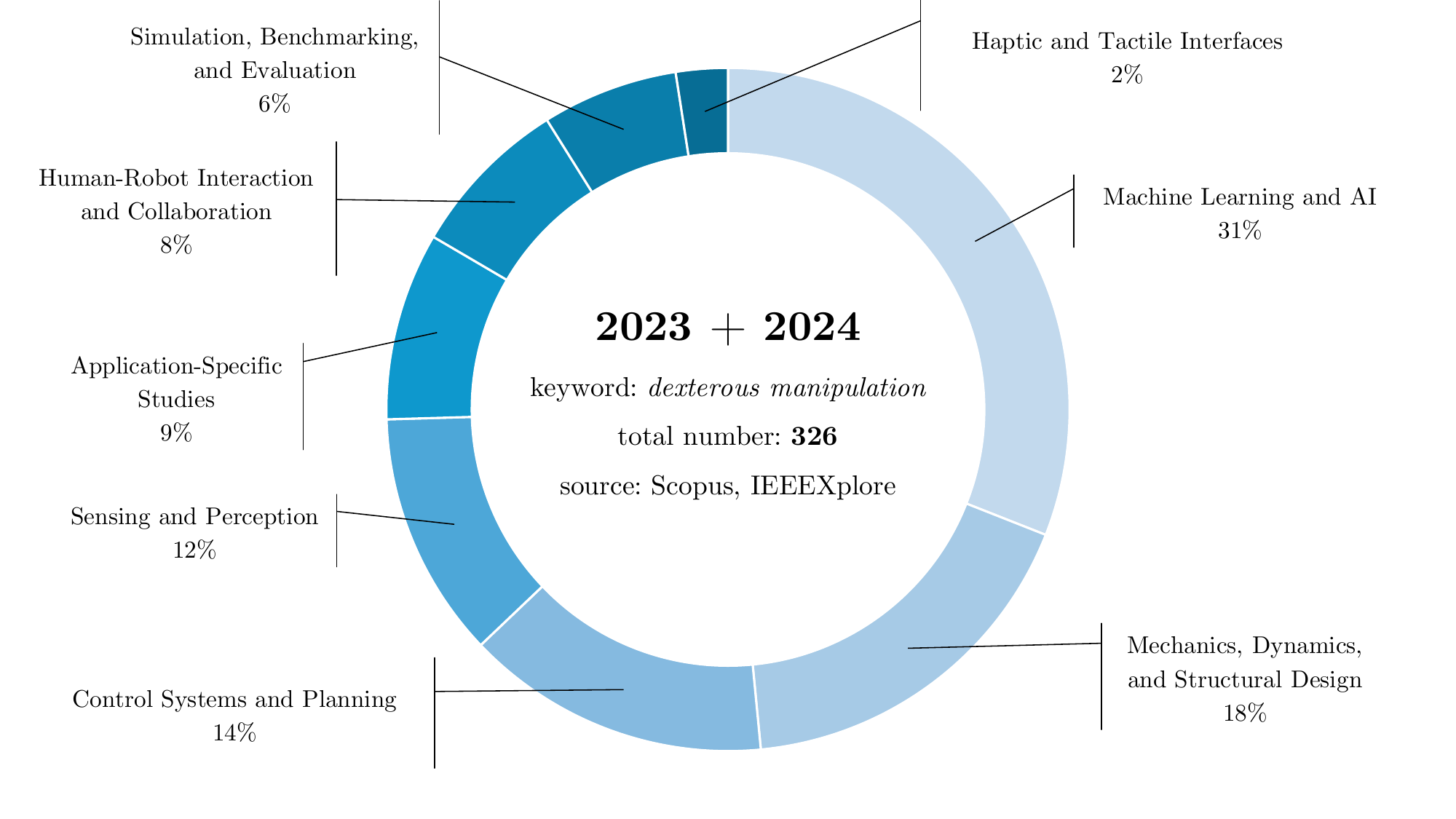}
    \caption{Distribution of publications from 2023 and 2024 with the keyword "dexterous manipulation" across different categories.}
    \label{fig:dex_man_2023_cat}
\end{figure}

A comprehensive analysis of the latest state-of-the-art publications was conducted to identify emerging research trends in dexterous manipulation. 
We summarized the publications into the eight distinct research categories, listed in Table \ref{tab:dex_man_2023_cat}. Figure \ref{fig:dex_man_2023_cat} shows the distribution across these categories. Research in dexterous manipulation primarily focuses on machine learning algorithms and the design of dexterous manipulators. Notably, haptic and tactile interfaces are not represented in large numbers, although they offer, in our opinion, great potential for interactive learning between humans and robots, as tactile feedback to the human teacher is crucial to performing fine motor tasks and perceiving an object's shape, size, texture, and weight \citep{Jin2023Progressinteractions, Dahiya2010TactileHumanoids}.

\begin{table}[ht]
    \centering
    \caption{Categories in dexterous manipulation research and their focus areas.}
    \begin{tabular}{@{}lp{8cm}@{}}
        \toprule
        \textbf{Category} & \textbf{Focus} \\
        \midrule
        Machine Learning and AI & Integrating advanced AI techniques such as reinforcement learning, deep learning, and machine learning frameworks to enhance robotic dexterity.\vspace{0.2cm} \\
        
        Sensing and Perception & Developing and implementing tactile sensors, cameras, and other sensory technologies to improve dexterous manipulation.\vspace{0.1cm} \\
        
        Control Systems and Planning & Control algorithms, motion planning, and optimization techniques to improve the accuracy and reliability of dexterous manipulation.\vspace{0.1cm} \\
        
        Human-Robot Interaction and Collaboration & Enhancing the interaction between humans and robots, including teleoperation, collaborative robots, and human-like dexterity.\vspace{0.1cm} \\
        
        Mechanics, Dynamics, and Structural Design & Design, analysis, and modeling of robotic mechanisms, including actuators, soft robotics, and multi-fingered grippers.\vspace{0.1cm} \\
        
        Application-Specific Studies & Specific applications of dexterous manipulation in various fields such as surgery, industrial automation, and hazardous environments.\vspace{0.1cm} \\
        
        Simulation, Benchmarking, and Evaluation & Simulation, benchmarking, and performance evaluation to assess and improve dexterous manipulation techniques.\vspace{0.1cm} \\
        
        Haptic and Tactile Interfaces & Development and application of haptic and tactile feedback systems to improve the realism and precision of robotic manipulation.\\
        \bottomrule
    \end{tabular}
    \label{tab:dex_man_2023_cat}
\end{table}

The category of machine learning algorithms for dexterous manipulation primarily includes grasp synthesis and manipulation skill/policy learning approaches. Grasp generation approaches utilize mainly classical neural networks \citep{Bla2023DMFC-GraspNetScenes}, Conditional VAriational Auto-Encoder (CVAE) \citep{Zhao2024GrainGraspGuidance}, normalizing flow \citep{Feng2024FFHFlowTime} or reinforcement learning \citep{Osa2018AnLearning}. Cluttered scenes represent one clear challenge in grasp generation \citep{Bla2023DMFC-GraspNetScenes}, where grasping planning also needs to consider other objects and uncertain observations, e.g., occlusions or partial observations of objects \citep{Chen2024SpringGraspUncertainty, Hidalgo-Carvajal2023AnthropomorphicCompletion, Farias2024Task-InformedObjects}. Policy learning for dexterous manipulation is dominated by reinforcement learning and imitation learning approaches \citep{Li2024ContinualManipulation, Wang2024Multi-StageManipulation, Han2024LearningObservations, Ze2023H-InDexManipulation, Li2023Within-HandLearning}. 
Human-in-the-loop approaches are particularly beneficial in policy learning, as this domain involves long-term decision-making where interactive feedback can significantly shape behavior over time \citep{Liu2024Robotdeployment, Chisari2022CorrectManipulation}. In contrast, grasp generation is typically a single-step task, making it less suited for effective human guidance. Therefore, this survey focuses on policy learning in real-world environments, where human input can have the most significant impact.

\section{Real-world Learning for Dexterous Manipulation}

Real-world learning for robots is challenging due to the high cost of training data, especially for high \ac{DoF} robots and tasks. Integrating human prior knowledge can accelerate autonomous robot learning \citep{Rayyes2023Interest-DrivenRobots, Rayyes2021EfficientRobots}. We will survey how previous work has dealt with dexterous manipulation using imitation learning, reinforcement learning, and \acl{IIL}.  

\subsection{Imitation Learning}

Imitation learning has emerged as a powerful tool in dexterous manipulation, enabling robotic systems to perform complex tasks by learning from human demonstrations. The versatility of imitation learning is showcased in a wide array of applications where robots are required to replicate human movements. However, the applications of dexterous manipulation using imitation learning are currently confined to relatively simple tasks. For instance, in a work by Amor in 2012, the focus was on the grasping of different mugs \citep{Amor2012Generalizationhands}. A more recent investigation by Ruppel expanded the scope to include pick-and-place operations, wiping tasks, and opening bottles using the Shadow Dexterous Hand \citep{Ruppel2020LearningDemonstration}. Yi evaluated the grasping abilities of an Allegro Hand, testing it on ten different objects in simulation and five objects in a real-world environment \citep{Yi2022AnthropomorphicLearning}. Arunachalam's work, titled DIME, explored manipulation tasks such as flipping a rectangular object, spinning a valve, and rotating a cube on the palm using an Allegro Hand \citep{Arunachalam2023DexterousManipulation}. Moreover, the Holo-Dex extension incorporated tasks like card sliding and can spinning \citep{Arunachalam2023Holo-DexReality}. Publications that deal with long-horizon tasks specifically for dexterous manipulation are very rare. DexSkills is an exception \citep{Mao2024DexSkillsTasks}. It supports the hierarchical construction of long-horizon tasks composed of primitive skills. In experiments, 20 primitive skills were used to create and execute various long-horizon tasks. For example, lifting and moving a box object with the Allegro Hand \citep{Mao2024DexSkillsTasks}. Recent work with diffusion policies performed experiments on more complex manipulation tasks like wrapping plasticine, making dumpling pleats, and pouring \citep{Ze20243DRepresentations}. However, all those applications do not fully utilize the dexterity of an anthropomorphic hand. There is still significant potential in leveraging the multi-dimensional capabilities of robot hands for complex human-like manipulation, including using tools made for humans to extend the field of application. 

While imitation learning has demonstrated promising results in dexterous manipulation, these achievements are closely tied to the specific methods employed. To understand how such outcomes are realized, it is essential to examine the diverse approaches developed to overcome the key challenges inherent to this domain. The following sections present a range of works that address three central challenges posed by dexterous manipulation: managing high-dimensional action spaces, handling multi-modal contact interactions, and enabling long-horizon task execution.

\subsubsection{The high-dimensional action space:} The biggest challenge in dexterous manipulation lies in learning a policy capable of managing the high complexity required to control such a large action space effectively. It is known from statistical learning that the amount of training data generally increases with the complexity of the model \citep{Hastie2009TheLearning, Sutton2018ReinforcementIntroduction, Lu2024OvercomingFactorization, Kubus2018LearningEfficiently}. It is, therefore, crucial to select a sample-efficient method to learn the policy with a manageable amount of data. However, this results in a trade-off with the policy's reduced generalizability and dexterity, limiting its applications. Potential sample-efficient methods include splitting the policy into a learned and hard-coded part \citep{Yi2022AnthropomorphicLearning}, reducing the action space to a low-dimensional latent space \citep{Amor2012Generalizationhands, Hu2022LearnDemonstration, Liconti2024LeveragingHand, He2022DiscoveringLearning}, reducing the input dimensionality of the policy in visual imitation learning \citep{Ruppel2020LearningDemonstration, Cai2024VisualRearrangement}, or using non-parametric policies without learnable parameters \citep{Arunachalam2023DexterousManipulation, Arunachalam2023Holo-DexReality}. 

An example of splitting the policy into two parts was presented in \citet{Yi2022AnthropomorphicLearning} for grasping different objects. One part of the policy takes care of reaching the objects. As this part only considers the movement of the robot arm, it is fundamentally not different from a task with a non-dexterous gripper. Therefore, classical methods like \ac{DMP} can learn this part of the policy efficiently with a small number of demonstrations \citep{Saveriano2023DynamicSurvey}. The second part consists of the actual grasping of an object, which includes identifying the object, predicting the object pose, and selecting a predefined grasping pose for the fingers. The finger motion planning is executed through a predefined strategy, as learning these finger trajectories via \ac{DMP} has resulted in poor performance due to the high dimensionality. 

Reducing the action space into a low-dimensional latent grasp space was inspired by how humans control their hands. Research studies show that the individual muscles in the hand are not controlled individually. Instead, the fingers are controlled by hand synergies \citep{Santello2016Handhands, Starke2024KinematicGeneration, He2022DiscoveringLearning}. These hand synergies can be modeled as a projection of the configuration space of the hand into a low-dimensional space. For instance, \citet{Amor2012Generalizationhands} utilizes the principal component analysis (PCA) for this projection, where the first component corresponds to the opening and closing of the hand, and higher-order components are used for more detailed hand motions. \citet{Amor2012Generalizationhands} states that only five dimensions are required to represent the relevant grasp movements. Using this low-dimensional grasp space, a \ac{DMP} can be used to define the policy for the finger movement and a separate \ac{DMP} for the wrist pose. The policy output is then mapped back to the original high-dimensional action space afterward. While PCA represents a linear mapping, a Variational Autoencoder (VAE) can learn a more complex mapping from the high-dimensional action space to a low-dimensional latent space from task-agnostic datasets, reducing the amount of expensive, task-specific training data in \ac{BC} \citep{Liconti2024LeveragingHand}.

Rather than simplifying the action space, some approaches focus on reducing the dimensionality of the policy's input. This is particularly relevant—and increasingly common—in visual imitation learning, where policies are conditioned on high-dimensional image data \citep{Nair2022R3MManipulation}. While not exclusive to dexterous manipulation, these techniques play a significant role in making policy learning more tractable in visually rich environments \citep{Liu2025VTDexManiplearning}. In \citet{Ruppel2020LearningDemonstration}, the manipulated objects and the robot hand are represented as point sets. Feed-forward and recurrent policy networks are trained using this representation, which is constructed either by manually attaching LEDs to track positions or by using a CNN to generate virtual keypoints. Similarly, \citet{Gao2023K-VILLearning} and \citet{Gao2024Bi-KVILTasks} use visual keypoints and geometric constraints to learn movement primitives, forming an object-centric task representation. The Multifeature Implicit Model (MIMO) \citep{Cai2024VisualRearrangement} introduces a novel object representation that incorporates multiple spatial features between a point and an object. This approach enhances the performance of visual imitation learning for task-oriented object grasping with a robot hand.

While the previous methods use a parameterized policy in the form of \ac{DMP} or neural networks, whose weights must be learned, non-parametric approaches derive the action to be executed directly from the training data. No time-intensive training on the policy is required. DIME \citep{Arunachalam2023DexterousManipulation} is an example. Here, state-based or image-based observations are utilized to find matches in the demonstrations via the nearest neighbor method to extract the following action. For image-based observations, dimensionality reduction with Bootstrap Your Own Latent (BYOL) \citep{Grill2020BootstrapLearning} is performed before applying the nearest neighbor algorithm in the embedding space to find the action. 

Still, several challenges and limitations hinder the current approaches to imitation learning for dexterous manipulation. When representing objects as point sets, only relying on a limited number of point markers from a motion tracking system as observations restricts the agent from supporting higher-dimensional observations like point clouds or images, which are necessary for more precise manipulation tasks \citep{Ruppel2020LearningDemonstration}. Relying on predefined grasps for each object limits the system's effectiveness when dealing with unknown objects \citep{Yi2022AnthropomorphicLearning}, which is essential when expanding the application scope from a structured environment to an open world, such as households. Non-parametric approaches suffer from low success rates and low generalization in task situations where the visual complexity of the input can not be adequately encoded in latent space \citep{Arunachalam2023DexterousManipulation, Arunachalam2023Holo-DexReality}. Enlarging the latent space, on the other hand, requires more data to learn the encoder. The dimensionality reduction of the action space via principal component analysis results in low generalization to new tasks \citep{Amor2012Generalizationhands}. Due to the information loss, the dexterity of hand motions is reduced.

\subsubsection{Multi-modality due to contact interactions:} Dexterous manipulation tasks often involve contact-rich interactions with the environment, leading to inherently multimodal action distributions—for example, multiple valid grasps or contact sequences that achieve the same goal. Naively averaging over such demonstrations, as in standard behavior cloning, often yields unnatural or ineffective actions. To address this, the key idea is to use probabilistic or structured policies that can represent a distribution over actions rather than a single deterministic output. Over the past few years, multiple solutions and combinations of those have been presented, primarily for non-dexterous tasks \citep{Urain2024DeepDemonstrations}:
\begin{itemize}
    \item Latent variable / Sampling Models—such as VAEs, GANs, and Normalizing Flows—address multi-modality by explicitly modeling distributions over actions through a latent space conditioned on context. These models enable efficient sample generation and capture diverse behavioral modes, making them well-suited for representing the inherent variability in contact-rich manipulation tasks. One popular example of this category is the Action Chunking with Transformer (ACT) algorithm \citep{Zhao2023LearningHardware}.
    \item Mixture Density Models (MDMs) represent action distributions as weighted combinations of parametric densities—typically Gaussians—conditioned on contextual inputs. By modeling multiple modes explicitly, MDMs provide a principled approach to capturing multi-modality in continuous action spaces \citep{Shafiullah2022Behaviorstone, Zhu2023VIOLAPriors, Mees2022WhatData}.
    \item Energy-Based Models (EBMs) define action distributions implicitly via an energy function, where low-energy regions correspond to high-probability behaviors. Sampling typically requires iterative optimization or stochastic methods such as Langevin dynamics \citep{Florence2022ImplicitCloning}.
    \item Discretized Action Models / Categorical models approximate continuous action spaces by discretizing them into a finite set of tokens, enabling the use of classification-based generative architectures. These models capture multi-modality by representing action distributions as categorical probabilities over discrete tokens, often leveraging spatial value maps or autoregressive structures to model complex, high-dimensional behaviors \citep{Shafiullah2022Behaviorstone, Brohan2023RT-1Scale, Zitkovich2023RT-2Control}.
    \item Diffusion Models (DMs) generate samples through an iterative denoising process that transforms noise into structured data, effectively modeling complex distributions over actions. By parameterizing the score function of an implicit energy landscape, DMs capture multi-modality via successive refinements of noisy inputs, enabling expressive and composable generative modeling despite slower inference compared to direct sampling approaches \citep{Chi2024DiffusionDiffusion, Reuss2024MultimodalGoals, Wolf2025DiffusionSurvey, Freiberg2025DiffusionGrasping}.
\end{itemize}

Diffusion models, in particular, have gained popularity for modeling multimodal distributions in high-dimensional action spaces, making them well-suited for dexterous tasks requiring fine-grained contact reasoning. Their ability to generate diverse behaviors that reflect real-world stochasticity comes at the cost of increased computational demands and reduced sample efficiency. For example, state-of-the-art diffusion policies often require over 100 expert demonstrations to achieve proficiency \citep{Chi2024DiffusionDiffusion, Chi2024UniversalRobots, Pearce2023ImitatingModels}. An exception is the 3D Diffusion Policy (3DP) \citep{Ze20243DRepresentations}, which enhances training efficiency by encoding sparse point clouds into compact 3D representations using a lightweight MLP encoder. Conditioning on these compact representations—rather than raw sensory inputs—accelerates learning and improves generalization, enabling successful policy training with as few as 10–40 demonstrations across both simulated and real-world dexterous manipulation tasks. Notably, omitting color information from point clouds further increases robustness to novel objects. Building on 3DP, FlowPolicy \citep{Zhang2025FlowPolicyManipulation} introduces an extension that significantly improves inference speed. By employing consistency flow matching, it enables action generation from noise in a single inference step, while maintaining comparable success rates.

The diversity of approaches for modeling multi-modal action distributions reflects the complexity of contact-rich manipulation tasks. Each class of policies offers distinct trade-offs. Latent variable and mixture density policies are lightweight and support efficient sampling, but may struggle with highly complex or discontinuous behaviors. Energy-based and diffusion policies provide greater expressiveness and compositionality but incur slower inference and higher data requirements. Diffusion methods can be viewed as conceptually bridging EBMs and latent-variable approaches: they exploit score-based training to navigate implicit energy landscapes while maintaining a structured generative process. Yet their computational demands and reliance on large datasets limit applicability in low-data or real-time settings. Discretized action policies provide a pragmatic alternative by converting continuous control into sequence prediction, allowing the use of powerful architectures like transformers. Their performance, however, depends on discretization granularity and can suffer from reduced precision in fine motor tasks.

\subsubsection{Performing long-horizon tasks:} Policies for long-horizon tasks are primarily formulated at a higher level of abstraction and do not consider the low-level control of individual joints. This area of research is often referred to as task planning \citep{Russell2016Artificialapproach, Guo2023RecentSurvey}. Here, imitation learning is also popular in learning task plans from human teachers \citep{Diehl2021AutomatedObservations, Ramirez-Amaro2017Transferringactivities}. Although research in dexterous manipulation mainly focuses on low-level policies for short-horizon tasks, some approaches combine both levels, such as DexSkills \citep{Mao2024DexSkillsTasks}. DexSkills views a robot's task from a hierarchical perspective to be able to execute long-horizon tasks. The core idea is to decompose complex tasks into primitive skills. The system includes several key components. First, a temporal autoencoder extracts a latent feature space from the demonstrations. Instead of images, the observation consists of the robot's joint states, tactile information, and contact status. Second, a label decoder segments a task into primitive skills based on the latent representation. Finally, a multilayer perceptron (MLP) learns the state-action pairs for each primitive skill separately from human demonstrations via \ac{BC}. After the initial training, new long-horizon tasks can be learned from a single demonstration. The primitive skill sequence is extracted via the label decoder and autonomously executed by the robot using the provided label sequence. Although DexSkills is promising, its' reliance on predefined primitive skills limits the range of possible long-horizon tasks. Additionally, segmenting demonstrations into individual primitive skills requires supervised training with labeled data, making data acquisition expensive despite the relatively small amount of training data needed \citep{Mao2024DexSkillsTasks}. 

To address these challenges, recent research has begun exploring the integration of Vision-Language Models (VLMs) into imitation learning pipelines. Approaches such as RoboDexVLM \citep{Liu2025RoboDexVLMManipulation} and DexGraspVLA \citep{Zhong2025DexGraspVLAGrasping} leverage the generalization capabilities of large-scale pretrained models to interpret high-level task descriptions and infer action sequences from diverse visual inputs, including third-person videos. These models offer a way to bridge the embodiment and viewpoint gap by grounding language in visual context and enabling robots to reason about tasks in a more abstract, flexible manner. By combining the structured decomposition of tasks—as seen in hierarchical methods like DexSkills—with the broad priors and multimodal understanding of VLMs, these approaches promise more scalable and generalizable solutions for long-horizon dexterous manipulation. But with the use of VLMs for task planning and reasoning, the inference time increases significantly, which limits real-time deployment. Additionally, the predefined primitive skills impose a ceiling on the generalization capabilities of such methods.

Another complementary direction is leveraging large-scale human demonstration videos—especially those sourced from the internet—as a means to improve data diversity and reduce the cost and effort of expert data collection \citep{Shaw2023VideoDexVideos, Shaw2024Learningvideos, Mandikal2022DexVIPVideo, Sivakumar2022RoboticYoutube, Qin2022DexMVVideos}. Such videos offer a rich prior over manipulation behaviors and object interactions, which can support better generalization across tasks and environments. However, bridging the domain gap between third-person human video and robot execution, particularly with dexterous hands, introduces substantial challenges. These include differences in embodiment (e.g., human hands vs. robot hands), viewpoint and occlusion issues, the lack of action labels, and the difficulty of inferring precise 3D contact-rich motions from 2D videos.

The limitations discussed above underscore the need for more robust and flexible approaches to advance imitation learning in dexterous manipulation. The highlighted publications reveal two most prominent challenges: generalization across tasks and the ability to handle long-horizon behaviors. Methods such as DexSkills, vision-language models (VLMs), and 3D diffusion policies offer promising directions to address these issues. A particularly compelling avenue for future research lies in combining hierarchical skill composition, as demonstrated in DexSkills, with the generative flexibility of diffusion policies. Such integration could reduce overall training effort by enabling the reuse of learned skills across diverse tasks, thereby improving scalability and efficiency.

Still, with imitation learning approaches, major limitations remain. In addition to the covariant shift problem, imitation learning approaches that have learned behavior in a supervised mode have the disadvantage that the quality of the demonstrations limits the performance of the resulting policy. Consequently, demonstrations represent an upper bound that the policy cannot exceed. To improve the policy beyond the level of the demonstrations, a criterion must be defined that can be used to measure and optimize performance. This problem is addressed in reinforcement learning, in which a reward function is defined that provides the agent with feedback on its performance. Based on this feedback, the agent can optimize its behavior and thus improve its performance. In the context of reinforcement learning, a policy is learned by the agent through numerous interactions with the environment. The data collected in the process is used to train the policy. The requirement for extensive interaction poses a significant challenge when training a robot using reinforcement learning in the real world. Firstly, the large number of trials can be very time-consuming, and human intervention is required to reset the environment between episodes. Secondly, the actions the agent selects may be hazardous or cause physical damage to the hardware, posing risks to both the robot and its surroundings and limiting their real-world learning. However, some approaches allow exploiting the potential of reinforcement learning on physical robots, including dexterous manipulation. These will be discussed in more detail in the next section.

\subsection{Reinforcement Learning}

Using reinforcement learning for real-world dexterous manipulation is categorized in three directions in the literature \citep{Yu2022DexterousReview}.
The classical reinforcement learning approach involves learning from scratch, in which a novice agent learns purely by interacting with a physical robot and the environment. The second approach uses a pre-trained agent to begin learning from a safer and more informed initial policy, typically achieved by combining reinforcement learning with imitation learning. Lastly, the agent is trained in simulation, and then the policy is transferred to the physical robot. Some relevant examples of those approaches are presented below.

\subsubsection{Reinforcement Learning from scratch:} Learning from scratch is mainly done in simulation, as data generation is significantly more cost-effective. Implementing this approach directly on a physical robot poses additional challenges beyond those already mentioned, particularly the difficulty of accurately retrieving the representation of the environment's state. Perceiving the environment is typically achieved using vision-based sensors, such as cameras \citep{Haarnoja2018SoftApplications, Luo2021RobustStudy}. However, this approach comes with a trade-off in sample efficiency, as the high-dimensional data generated increases both computational demands and processing complexity.
Instead, introducing tactile sensors in observation reduces the sample complexity for dexterous manipulation significantly \citep{Melnik2019TactileTasks}. \citet{Hoof2015Learningfeatures} uses pure tactile information while \citet{Falco2018OnManipulation} combines tactile and visual observations in policy learning with reinforcement learning for an in-hand manipulation task. To avoid costly resets of the environment by humans, \citet{Gupta2021Reset-FreeIntervention} presents a reset-free approach, where learning multiple tasks simultaneously and sequencing them solves the problem automatically. The individual tasks provide the reset for other ones.

\subsubsection{Reinforcement Learning and Demonstrations:} Combining reinforcement learning with imitation learning overcomes issues of both approaches \citep{Nair2018OvercomingDemonstrations, Hester2018DeepDemonstrations}. On the one hand, the sample complexity is significantly reduced by using demonstrations to pre-train a policy. On the other hand, through reinforcement learning, the robot can obtain further information through interaction with the environment and fine-tune its performance beyond the quality of demonstrations. Nevertheless, especially in the field of dexterous manipulation, many approaches still rely on training and experiments in simulation due to efficiency, safety, and cost reasons \citep{Rajeswaran2018LearningDemonstrations, Radosavovic2021State-OnlyManipulation, Huang2023Dexteroushand, Mosbach2022AcceleratingDemonstrations, Qin2022DexMVVideos, Han2024LearningObservations, Mandikal2022DexVIPVideo, Orbik2021Inversemanipulation}. Only a few studies use reinforcement learning with demonstrations on physical robots \citep{Gupta2016Learningdemonstrations, Zhu2019DexterousLow-Cost, Nair2020AWACDatasets}. Their ability to conduct real-world experiments hinges on the constraint of basing their policy on low-dimensional state spaces, resulting in a lower-capacity policy that can be efficiently trained on less data.

Utilizing object-centric demonstrations—focusing solely on the manipulated object's trajectory—has shown potential in training tasks for soft robotic hands, as illustrated in \citet{Gupta2016Learningdemonstrations}. Their approach is based on Guided Policy Search (GPS), which offers advantages in learning high-dimensional tasks. Multiple policies are initially learned from demonstrations and refined using model-based reinforcement learning to follow the demonstrated behavior closely. These are then distilled into a single neural network policy via supervised learning. However, the final policy's performance was limited, primarily due to the soft and compliant nature of the RBO Hand 2 robotic hand. More effective results have been achieved with the Demo Augmented Policy Gradient (DAPG) method introduced by \citet{Rajeswaran2018LearningDemonstrations}, which has become a widely adopted approach in dexterous manipulation \citep{Qin2022DexMVVideos, Huang2023Dexteroushand, Qin2022FromTeleoperation}. It combines model-free, on-policy reinforcement learning with imitation learning to enhance policy exploration and reduce sample complexity. Pre-training with \ac{BC} equips the agent with an intuitive understanding of task-solving strategies before its autonomous exploration. Although initially tested in a simulated environment, DAPG’s applicability to real-world manipulation was later validated by \citet{Zhu2019DexterousLow-Cost} using an Allegro Hand. Demonstrations were shown to significantly accelerate training in the real world — cutting training time from 4-7 hours without demonstrations to 2-3 hours with 20 demonstrations. Another approach, the Advantage-Weighted Actor-Critic (AWAC) algorithm introduced by \citet{Nair2020AWACDatasets}, combines offline reinforcement learning with online fine-tuning using off-policy methods. Off-policy algorithms are generally more sample-efficient, as they can reuse offline data during the online learning phase. AWAC formulates the policy improvement step as a constrained optimization problem, ensuring that the updated policy remains close to the expert demonstrations. Its effectiveness is demonstrated on an object-repositioning task using a four-fingered dexterous robotic hand. The authors also note that tuning AWAC’s parameters can be challenging. A direct comparison between DAPG and AWAC on dexterous manipulation tasks shows that AWAC can achieve faster learning and better data efficiency than DAPG \citep{Nair2020AWACDatasets}.  

\subsubsection{Sim-to-real Reinforcement Learning:} 
While some studies have successfully implemented reinforcement learning directly on physical robots, with or without demonstration, simulation remains the most practical and widely utilized environment for training due to its cost-effectiveness, scalability, and safety. 
The complete freedom concerning safety restrictions and the almost unlimited data availability through parallelization offer a unique starting point for reinforcement learning. Still, sim-to-real zero-shot achieves only limited performance due to the reality gap \citep{Gilles2024MetaGraspNetV2Grasping, Gilles2025MetaMVUCGrasping}. 
Especially for dexterous manipulation with its complex dynamics, it is challenging to create a simulation model that corresponds to reality. Developing more powerful and realistic simulators like Isaac Sim\footnote{\url{https://developer.nvidia.com/isaac/sim}}, GENESIS\footnote{\url{https://genesis-embodied-ai.github.io/}}, and MuJoCo \citep{Todorov2012MuJoCocontrol} will not completely close the reality gap. A common method to overcome this gap is domain randomization, where the simulation is randomized with disturbances to compensate for inaccuracies in the modeling \citep{Kumar2019ContextualPolicies, Akkaya2019SolvingHand, Andrychowicz2020Learningmanipulation}. This can include the randomization of light, textures, or friction parameters. Randomization facilitates the agent's adaptation to a wide range of environments, where the real world might represent one instance of this spectrum. Training a single robotic hand to solve a Rubik's cube shows how powerful sim-to-real methods can be \citep{Akkaya2019SolvingHand}, but also how much computational resources are required to train such a complex policy: over 900 parallel workers were used over multiple months to collect data corresponding to 13.000 years of experience in simulation. We believe that using such a vast amount of computational resources to train a single task is not an effective approach. Instead, it may be more reasonable to leverage simulators and computational resources to pretrain a comprehensive foundation model with general knowledge — for example, about physical properties of the world — so that task-specific knowledge can subsequently be fine-tuned more efficiently. This was done for example by NVIDIA, when training the Generalist Robot 00 Technogly (GR00T) model on synthetic data from simulators and also real world data \citep{Bjorck2025GR00TRobots}.

While approaches based on reinforcement learning have shown impressive results in dexterous manipulation, particularly by utilizing sim-to-real methods for transferring policies from simulation to physical robots, they still face significant limitations. Data efficiency and safety remain key challenges, even when using demonstrations for a warm start. Additionally, the reliance on sim-to-real methods limits the system's ability to adapt effectively to unstructured and highly dynamic environments.

\subsection{Interactive Imitation Learning for Dexterous Manipulation}

As outlined in the introduction, integrating humans interactively into the learning process represents a promising research direction. This approach helps mitigate challenges in imitation learning, such as covariate shift, and reduces the effort required for collecting demonstrations. Despite its potential, only a limited number of works have explored interactive human involvement in real-world dexterous manipulation tasks \citep{Kaya2018EffectiveControl, Argall2011TactileAdaptation, Ugur2011Learningscaffolding, Sauser2012Iterativecorrections, Ding2023LearningPreference, Si2024TildeDeltaHand, Wang2024DexCapManipulation}.
Not all of these works aim to learn generalized policies. Some intersect with shared control schemes \citep{Kaya2018EffectiveControl}, while others focus on identifying graspable regions of objects \citep{Ugur2011Learningscaffolding}. Among those that do learn policies, the objectives vary: learning stable grasps \citep{Sauser2012Iterativecorrections}, grasping strategies \citep{Argall2011TactileAdaptation}, human-like motion characteristics \citep{Ding2023LearningPreference}, or complex dexterous manipulation skills \citep{Wang2024DexCapManipulation, Si2024TildeDeltaHand}.

The following section presents these works to provide an overview of the field and to highlight how they differ from the understanding of \acl{IIL} adopted in this survey. The limited number of publications suggests that \ac{IIL} has yet to gain widespread traction in dexterous manipulation, though recent contributions offer promising starting points.

In \citet{Kaya2018EffectiveControl}, an in-hand manipulation is learned interactively using a 16-DoF robotic hand. The task consists of swapping the position of two balls in the hand using skillful finger and hand movements. The human operator learns to control the robot arm via kinesthetic teaching while the hand runs a periodic finger movement. After the human operator has learned how to control the robot, successful executions of human and robot control are combined into a fully autonomous policy via regular imitation learning with a \ac{DMP}. Although the human operator interacts with the robot during execution, this is not interactive imitation learning; instead, the interactive part belongs to the shared control scheme \citep{Abbink2018ADiversity}, as the policy is learned after all data have been collected, like in normal imitation learning. 

In \citet{Ugur2011Learningscaffolding}, a 16-DoF robotic hand is used to grasp and lift various objects. The learning is supported by a human teacher and called parental scaffolding.  Initially, the robot performs rough reach motions towards the object's center. The human teacher can interfere by physical contact with the robot's motion to achieve successful grasping. The robot checks the distance between its fingers and the object and records "first-touch" points on the object in case of contact, which correspond to graspable parts of the objects. A classifier that differentiates whether a voxel is graspable is trained using these "first-touch" points. The classifier is based on a newly proposed metric that captures the relationship between graspable voxels and all voxels of the object. The main focus lies on learning and inferring graspable parts of the objects, not the motion of grasping the object itself. Therefore, only a simple lookup table-based mechanism is used to select a reach-grasp-lift execution trajectory to grasp an object.

An early interactive imitation learning related work that learns a manipulation policy is presented in \citet{Argall2011TactileAdaptation}. It introduces the Tactile Policy Correction (TPC) algorithm to learn how to grasp simple objects with a dexterous 8-DoF robotic hand from human interactions. The approach consists of two phases. First, a dataset of human demonstrations is created through teleoperation, allowing the robot agent to derive an initial policy via \ac{BC}. In the second phase, the agent executes the policy and receives corrective tactile feedback from the human teacher to adapt the policy. The teacher physically touches the robot on five touchpads on the arm to provide feedback. Policy execution consists of object pose prediction and action selection. It utilizes \acs{GMM}-\acs{GMR}: \ac{GMM} encodes demonstrations, and \ac{GMR} predicts target poses. The \ac{GMM} is trained with weighted \ac{EM}. In a subprocess, an inverse kinematic controller is responsible for selecting the appropriate action to reach the target pose. The policy undergoes adaptation through recurrent derivation from the updated dataset. This dataset evolves through tactile corrections from the human teacher, either through policy reuse, involving modification of existing data points, or refinement, entailing the addition of new data points. The presented experiments show that refinement is more effective than providing more demonstrations via teleoperation.
A very similar approach is used to learn stable grasps for dexterous robot hands \citep{Sauser2012Iterativecorrections}. The grasping task is also modeled as \ac{GMM}; the difference is in how the human interacts with the robot. The human teacher provides corrections by pressing on the fingertips to encourage better contact and shift the pose as much as possible within the compliance of the hand. The robot generates self-demonstrations by following the pose-pressure pair recorded from the previous step. This process extends the dataset by incorporating data that is free from the influence of correction forces, enabling more accurate and autonomous learning.

A different kind of human feedback is used in \citet{Ding2023LearningPreference} to train a policy for a dexterous hand. Instead of showing how to do the tasks, the feedback of the human evaluates the policy. The policy is learned via reinforcement learning in simulation and later fine-tuned with human feedback to enhance its human-like characteristics. Although no demonstrations are used to train the policy, and only human feedback in the form of preferences over generated trajectories is provided, this approach still exemplifies interactive learning for dexterous manipulation. The human feedback primarily aims to make the policy execution more human-like. The human teacher is presented with two generated trajectories, and the provided feedback is the choice of which trajectory is more human-like. Based on this feedback, a reward model is trained to fine-tune the policy via reinforcement learning later. For execution on a physical robot, no additional training is performed; instead, the policy is directly transferred to the physical robot showing its robustness against the reality gap.

With the recent advent of diffusion models, diffusion-based imitation learning approaches have gained significant popularity. These models are particularly well-suited for managing multimodal action spaces in dexterous manipulation. In the work titled Tilde \citep{Si2024TildeDeltaHand}, a unique integration of diffusion-based imitation learning with DAgger-based on-policy updates is employed to perform dexterous manipulation tasks using a high \ac{DoF} robotic hand. This novel combination leverages the strengths of both methods, where diffusion models handle complex, multimodal action spaces effectively, and DAgger \citep{Ross2011ALearning} provides robust, real-time corrections through human teacher interventions in case of failures. The robotic hand, known as DeltaHand, features a non-anthropomorphic design with four fingers, each possessing 3 DoF. Demonstrations for seven distinct manipulation tasks are recorded using a kinematic twin teleoperation interface. A vision-conditioned diffusion policy is then learned, utilizing input from an in-hand camera and the joint states. Integrating on-policy expert corrections via DAgger helps mitigate covariate shifts, ensuring more reliable performance. However, the authors also note that generalization to unstructured environments is limited, and the movement of the robot arm for more complex tasks has not yet been considered. This is likely due to the direct conditioning of the policy on images and, therefore, the limited data used to train the policy. An approach to address this is DexCap \citep{Wang2024DexCapManipulation}, which introduces a portable hand motion capture system and an imitation learning framework. The motion capture system consists of a motion capture glove for finger tracking and three cameras for wrist tracking and environment perception. The portability allows the accessible collection of demonstration data for bimanual tasks. DexCap enables robots to learn bimanual dexterous manipulation from human motion capture data (i.e., human motion capture data) through a diffusion policy, using a point-cloud-based \ac{BC} algorithm. The learning process consists of three steps: First, the motion capture data is retargeted into the robot's operational space, which includes mapping the finger positions and 6-DOF wrist pose into the action space of the robot via inverse kinematics. Secondly, a diffusion policy is trained based on down-sampled colored point cloud and retargeted data. Using colored point clouds transformed into a consistent world frame as input for the policy, rather than RGBD images, the system maintains stable observations even when the camera moves. Lastly, human operators can intervene and provide on-policy corrections to correct unexpected robot behavior. Corrections can be supplied as residual actions on top of the policy's actions or by taking complete control and guiding the robot via teleoperation. Corrections and original demonstrations are used together to refine the policy. Experiments demonstrate a 33\% improvement by fine-tuning with corrections on six household manipulation tasks with two 16-\ac{DoF} robotic hands.

The works presented in this section are among the few that incorporate interactive human correction into the learning process for dexterous manipulation. The limited number of such studies may be attributed to several factors. One major constraint is the need for physical hardware to enable safe and effective human-robot interaction and to evaluate learning outcomes—an especially costly requirement in the context of dexterous manipulation. Additionally, the availability of suitable robotic hands remains limited, further restricting experimentation.
Historically, research in this domain has focused heavily on reinforcement learning (RL) as the primary method for solving complex tasks. However, the practical limitations of RL—particularly its inefficiency and resource demands when applied to real-world hardware—have prompted a shift in perspective. As these limitations became more apparent, interest in interactive imitation learning (\ac{IIL}) began to grow. In dexterous manipulation specifically, sim-to-real transfer remains a significant challenge, often requiring substantial engineering effort to bridge the gap between simulation and physical deployment.
More recently, the rise of humanoid robots in industry \citep{BMW2024, Reuters2024}, which are increasingly being used in roles traditionally filled by humans and are expected to operate human tools, has further fueled interest in efficient training methods. In this context, \ac{IIL} has emerged as a promising approach for improving learning efficiency in high-dimensional manipulation tasks. Despite its potential, the field remains largely unexplored. Current works demonstrate the viability of \ac{IIL}, but also reveal gaps—such as the limited use of tactile feedback and generalization in unstructured environments—that point to valuable directions for future research. Exploring how \ac{IIL} methods developed in other areas of robotic manipulation can be adapted to dexterous tasks may yield important insights and broaden the scope of innovation in this field.

\section{Interactive Imitation Learning in Robotics}

While this survey has focused on dexterous manipulation thus far, this chapter examines approaches to the role of interactive imitation learning in general robotics, aiming to identify its potential for dexterous manipulation. We categorize the approaches in \ac{IIL} into two directions, which differ in how humans provide real-time feedback to the agent. 

\textbf{1) Corrective feedback}, where the human teacher gives feedback on \textit{how to improve} the execution in the action domain. So either the human gives absolute actions that replace the agent's actions \citep{Kelly2019HG-DAggerExperts}, or the human provides relative corrections \citep{Celemin2019AnFeedback}, guiding the agent toward correct actions.

\textbf{2) Evaluative feedback}, where the human teacher gives feedback on \textit{how well} the agent performs \citep{Knox2008TAMERReinforcement}. In this case, the human teacher does not need task expertise, but only the ability to evaluate performance. This feedback can be delivered as absolute evaluative signals—commonly referred to as human reinforcement—or as relative preferences between different agent behaviors.

\begin{figure}[ht]
    \centering
    \includegraphics[width=\textwidth]{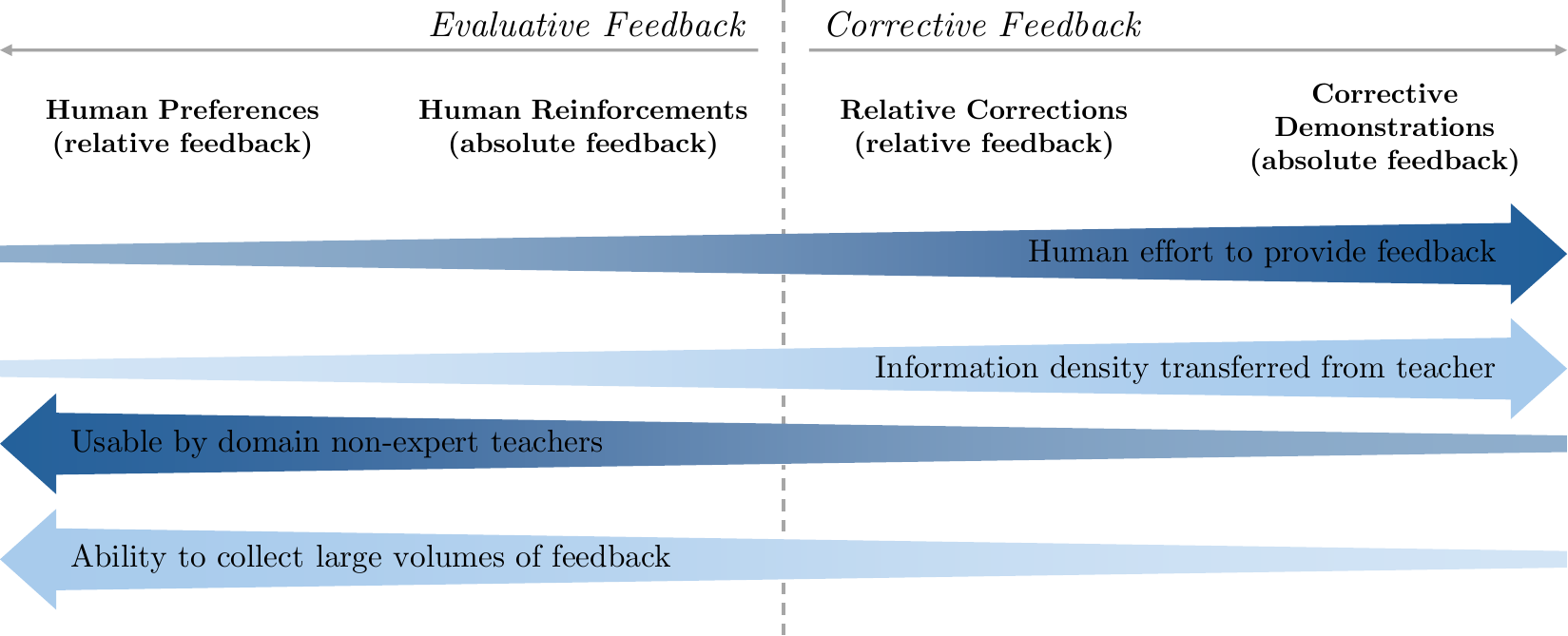}
    \caption{Comparison of Human Feedback Modalities, modified from \citet{Celemin2022InteractiveSurvey}}
    \label{fig:feedback_modalities}
\end{figure}

The choice of feedback modality in a learning application often depends on the level of autonomy desired in the agent’s exploration process. For example, evaluative feedback typically provides only limited guidance on the optimal policy, requiring the agent to rely more heavily on autonomous exploration. In contrast, corrective feedback or demonstrations convey more direct information about the policy, increasing reliance on human supervision. This trade-off is commonly described in the literature as the exploration-control spectrum \citep{Najar2021ReinforcementSurvey}. Figure~\ref{fig:feedback_modalities} summarizes common forms of human feedback, comparing them along key dimensions such as information content, human effort required, expertise needed, and scalability.

In IIL, various corrective feedback approaches have emerged from the original DAgger \citep{Ross2011ALearning} approach and developed further. While DAgger was one of the first algorithms to describe the interactive intervention of an expert in training, the expert in DAgger usually consisted of another algorithmic expert since this expert had to relabel each data sample of the agent. This would be too much work for a human expert, at least in robotics. Nevertheless, many approaches based on DAgger have been developed. For example, LazyDAgger \citep{Hoque2021LazyDAggerLearning}, where an additional learned meta-controller in the agent decides whether an expert should be consulted, thus reducing the number of human interactions. Instead of the agent deciding when it needs expert support, in most current approaches, the human teacher decides when a correction is necessary, for example, if the agent enters unsafe areas of the state space. HG-DAgger \citep{Kelly2019HG-DAggerExperts} is an algorithm where the human teacher (expert) chooses when to take over control. HG-DAgger is also used in the work RoboCopilot \citep{Wu2025RoboCopilotManipulation}, which presents a complete bi-manual teleoperation system that allows seamless human takeover by using a leader-follower approach. The teleoperation device is a kinematic replica of the robot arm with a user interface especially designed for an interactive learning setting.  While HG-DAgger involves only one expert who is as perfect as possible, MEGA-DAgger \citep{Sun2023MEGA-DAggerExperts} allows for several experts who may also be imperfect. A built-in filter resolves contrary corrections and removes unsafe demonstrations. HG-DAgger only learns from the generated data when the expert takes control, so-called supervisor-generated data. This means that HG-DAgger learns how to recover from error situations but not how to stay within the target area. In addition, the policy changes substantially in each iteration when training only with supervisor-generated data and ignoring agent-generated data. Intervention Weighted Regression (IWR) \citep{Mandlekar2020Human-in-the-LoopTeleoperation} addresses this problem by storing all the generated data. The agent-generated samples are stored in a separate dataset. Thus, one dataset has agent-generated data, and one has supervisor-generated data. During the policy training, an equal number of samples from both datasets are used to update the policy. This way, samples from interventions are weighted more heavily, with the idea that these samples are more likely to indicate bottlenecks in state space and thus be learned more robustly.

Other approaches integrate corrective feedback with reinforcement learning to leverage the efficiency of corrective feedback while benefiting from reinforcement learning's ability to optimize beyond experts' performance. For example, \citet{Celemin2019Reinforcementadvice} and \citet{Luo2021RobustStudy} propose learning a policy from supervisor-generated data (such as demonstrations and corrective feedback) while at the same time manually defining a reward function. This allows the agent to improve its policy continuously, achieving faster convergence than traditional reinforcement learning. However, shaping a reward function is a demanding engineering task. In another approach, \citet{Parnichkun2022ReILLearning} modifies the objective function of the reinforcement learning algorithm to incorporate the \ac{BC} objective, directly aligning it with supervisor-generated data but continuously improving its policy with reinforcement learning.

\ac{IIL}-algorithms utilizing evaluative feedback are a more direct way of bridging imitation and reinforcement learning. Many methods frame the robot control problem as a reinforcement learning task but impose the constraint that the environment lacks a predefined reward function. Instead, the agent's performance is assessed through evaluative feedback provided by a human teacher. An early example of this approach is the TAMER framework \citep{Knox2008TAMERReinforcement}, where a human teacher assigns scalar rewards based on its evaluation of the agent's behavior. The agent's objective is to select the action that maximizes this human-given reward for a given state. To achieve this, the agent learns a model of the human reward function using supervised learning techniques. \citet{Macglashan2017InteractiveFeedback} proposes an alternative approach in which human feedback depends on the agent's current policy maturity. He models human feedback as an advantage function, capturing the nuance that human teachers provide different evaluations depending on whether the agent is improving or performing adequately in the status quo. This more accurately reflects the dynamic nature of human feedback during the learning process. Although a reward function represents an evaluative feedback system in reinforcement learning, it can also be constructed by utilizing corrective feedback. For example, \citet{Luo2023RLIFLearning} and \citet{Kahn2021LaNDDisengagements} enable human teachers to give corrective feedback but only use it to indicate that an intervention occurred without considering the specifics of the intervention. These methods train an agent to minimize the probability of human intervention rather than focusing on the content of the feedback itself. This way, it's robust to imperfect experts, as the expert can also provide wrong corrections, but only the fact that a correction happens helps the agent. Another way to deal with humans lacking expertise in a task is through evaluative feedback in the form of preferences between two presented executions. On the exploration-control spectrum, this approach leans towards autonomous exploration, similar to classical reinforcement learning. However, as \citet{Christiano2017DeepPreferences} demonstrated, it can be applied to problems where a reward function cannot be explicitly defined. This method uses supervised learning techniques to infer a reward function from preference data. Since human preferences convey limited information, learning relies heavily on autonomous exploration, which often requires numerous and sometimes unsafe interactions between the agent and its environment. Consequently, this approach is primarily suited to simulated environments, making it less practical for real-world applications.

Combining corrective and evaluative feedback has produced promising results. For instance, in \citet{Spencer2020Learningfeedback}, in addition to recording corrected actions, each sample is tagged with a flag indicating whether the robot's current state is good enough. States without expert intervention are assumed to be acceptable. This state evaluation allows the agent to learn a value function in parallel, enabling it to refine its policy even without further input from the expert, as the current execution is generally considered good enough. However, human experts cannot provide corrections in every situation. The work "Correct me if I am wrong" \citep{Chisari2022CorrectManipulation} addresses this issue by extending IWR \citep{Mandlekar2020Human-in-the-LoopTeleoperation} to handle cases where no correction is possible. In such situations, evaluative feedback allows the expert to discard state-action pairs that might otherwise pollute the training data.

Not only the feedback type is a relevant categorization criterion for \ac{IIL} approaches, but also the different types of policy representations used are interesting. However, when examining different methods of policy representation, it becomes clear that the choice of representation, as in other robotic applications, depends heavily on the specific task rather than being unique to \ac{IIL}. A variety of function approximators for policy representation are used in \ac{IIL}, including linear models, radial basis functions (RBFs) \citep{Celemin2019AnFeedback}, classical feed-forward neural networks (FFNs) \citep{Kelly2019HG-DAggerExperts, Sun2023MEGA-DAggerExperts}, convolutional neural networks (CNNs) \citep{Pe2020InteractiveNetworks, Hoque2021LazyDAggerLearning}, recurrent neural networks (RNNs) / long short-term memory networks (LSTMs) \citep{Mandlekar2020Human-in-the-LoopTeleoperation, Chisari2022CorrectManipulation}, diffusion models \citep{Si2024TildeDeltaHand, Wang2024DexCapManipulation}, as well as \acp{DMP} \citep{Gams2016Adaptationinteraction, Celemin2019Reinforcementadvice} and Probabilistic Movement Primitive (ProMP) \citep{Ewerton2016Incrementalskills}. An overview is seen in Table \ref{tab:iil_works}.

\begin{table}[ht]
    \centering
    \caption{Overview of \ac{IIL} works based on policy representation and feedback type.}
    \begin{tabular}{@{}lll@{}}
    \toprule
     & \textbf{corrective feedback }  & \textbf{evaluative feedback}  \\ 
    \midrule
    linear models / RBFs & \makecell[l]{\citep{Celemin2019AnFeedback}, \\ \citep{Spencer2020Learningfeedback} }  & \makecell[l]{\citep{Knox2008TAMERReinforcement}, \\  \citep{Macglashan2017InteractiveFeedback}, \\ \citep{Spencer2020Learningfeedback}} \vspace{0.1cm}\\
    FFN & \makecell[l]{\citep{Kelly2019HG-DAggerExperts},\\ \citep{Sun2023MEGA-DAggerExperts}} &  \vspace{0.1cm}\\
    CNN & \makecell[l]{\citep{Pe2020InteractiveNetworks},\\   \citep{Hoque2021LazyDAggerLearning},\\ \citep{Luo2021RobustStudy},\\ \citep{Parnichkun2022ReILLearning}} & \makecell[l]{\citep{Luo2023RLIFLearning},\\ \citep{Christiano2017DeepPreferences}} \vspace{0.1cm}\\
    RNN / LSTM  & \makecell[l]{\citep{Mandlekar2020Human-in-the-LoopTeleoperation},\\   \citep{Parnichkun2022ReILLearning},\\ \citep{Chisari2022CorrectManipulation},\\ \citep{Wakabayashi2024Behavioralrate}, \citep{Liu2024Robotdeployment}} & \makecell[l]{\citep{Kahn2021LaNDDisengagements},\\ \citep{Chisari2022CorrectManipulation}} \vspace{0.1cm}\\
    Diffusion models  & \makecell[l]{\citep{Si2024TildeDeltaHand},\\ \citep{Wang2024DexCapManipulation}} &   \vspace{0.1cm}\\
    DMP / ProMP  & \makecell[l]{\citep{Gams2016Adaptationinteraction},\\ \citep{Celemin2019Reinforcementadvice},\\ \citep{Ewerton2016Incrementalskills}}  &   \\ 
    \bottomrule
    \end{tabular}
    \label{tab:iil_works}
\end{table}

The \ac{IIL} publications in this section clearly show that human feedback is integrated in many different ways. There is not only a rough distinction between evaluative and corrective feedback but also combinations of both variants, which have shown promising results. Transferring these algorithms to dexterous manipulation is an exciting area of research and presents unique challenges. For example, the question of how an interface between a robot and a human should be designed so that the human can comfortably provide valuable feedback to the robot. \ac{IIL} presents a promising avenue for advancing robotic capabilities, particularly in dexterous manipulation. Corrective feedback offers rich, intuitive, and safe guidance but necessitates experienced teachers and relies heavily on their performance. Evaluative feedback, while accessible to non-domain experts and easier to combine with recent trends of large language models, demands more data and interactions, posing potential safety risks. Including reward functions enables self-optimization, yet it comes with the high cost of reward engineering and inherent safety concerns due to autonomous exploration. Balancing these feedback mechanisms is crucial for developing robust and efficient \ac{IIL} systems adapting to complex real-world scenarios.

\section{Conclusion}

In this survey, we explored the current research landscape of \acl{IIL} (\ac{IIL}) for dexterous manipulation and identified a notable gap in concrete studies within this domain. We began by discussing the key challenges of dexterous manipulation, including its high-dimensional action space, multi-modal state representations, and long-horizon task complexity. We then presented various approaches for tackling real-world dexterous manipulation, such as imitation learning, reinforcement learning, and \ac{IIL}, highlighting both their potential and limitations. Given the scarcity of research explicitly addressing \ac{IIL} for dexterous manipulation, we extended our review to IIL methods used in other robotic applications, explaining how different approaches incorporate human feedback. 
\ac{IIL} for dexterous manipulation is an emerging field with substantial opportunities for further research. Notably, the use of on-policy corrections in \ac{IIL} has proven effective in mitigating covariate shift while enhancing sample efficiency, making it a promising direction for future advancements in dexterous robotic control. 
As the development of humanoid robotics continues to advance rapidly, the demand for efficient algorithms to equip these robots with the requisite skills is expected to increase significantly in the coming years. The growing interest among industrial companies in utilizing humanoid robots in production and logistics especially underscores the necessity for algorithms that can meet the industry's specific requirements, such as efficiency, flexibility, and real-world applicability.  This trend is further amplified by the recent surge in companies developing dexterous robotic hands, reflecting a broader shift toward enabling fine-grained manipulation capabilities in real-world settings. Consequently, these factors must be accorded a higher priority in developing algorithms than was previously the case in the research field of humanoid robots. In particular, \ac{IIL} approaches based on diffusion policies and those that use both corrective and evaluative feedback can provide a successful direction here. As task complexity increases, the importance of algorithms capable of handling long-horizon policies grows; thus, the integration of hierarchical approaches becomes crucial.

In summary, while \ac{IIL} for dexterous manipulation is still in its early stages, it holds great promise for enabling more efficient, scalable, and human-aligned robotic learning. Bridging the gap between current capabilities and real-world demands will require continued exploration of interactive learning paradigms, integration of tactile and multimodal feedback, and adaptation of successful strategies from broader robotic domains. This survey aims to serve as a foundation for future research and innovation in this exciting and rapidly evolving field.

\begin{acronym}

    \acro{IIL}{Interactive Imitation Learning}
    \acro{DMP}{Dynamic Movement Primitive}
    \acro{ProMP}{Probabilistic Movement Primitive}
    \acro{GMM}{Gaussian Mixture Model}
    \acro{GMR}{Gaussian Mixture Regression}
    \acro{EM}{Expectation-Maximization}
    \acro{BC}{Behavioral Cloning}
    \acro{DoF}{degrees of freedom}
    \acro{MLP}{multilayer perceptron}

\end{acronym}





\bibliographystyle{apalike} 
\bibliography{references}

\begin{thebibliography}{}

\bibitem[Abbink et~al., 2018]{Abbink2018ADiversity}
Abbink, D.~A., Carlson, T., Mulder, M., {de Winter}, J. C.~F., Aminravan, F.,
  Gibo, T.~L., and Boer, E.~R. (2018).
\newblock A topology of shared control systems---finding common ground in
  diversity.
\newblock {\em IEEE Transactions on Human-Machine Systems}, 48(5):509--525.

\bibitem[{AGILE ROBOTS}, 2023]{AgileHand}
{AGILE ROBOTS} (2023).
\newblock Agile hand {\textbar} agile robots se.
\newblock \url{https://www.agile-robots.com/en/robotic-solutions/agile-hand}.
\newblock Accessed on 2024-05-14.

\bibitem[Akkaya et~al., 2019]{Akkaya2019SolvingHand}
Akkaya, I., Andrychowicz, M., Chociej, M., Litwin, M., McGrew, B., Petron, A.,
  Paino, A., Plappert, M., Powell, G., Ribas, R., Schneider, J., Tezak, N.,
  Tworek, J., Welinder, P., Weng, L., Yuan, Q., Zaremba, W., and Zhang, L.
  (2019).
\newblock Solving rubik's cube with a robot hand.
\newblock {\em arXiv preprint arXiv:1910.07113}.

\bibitem[Andrychowicz et~al., 2020]{Andrychowicz2020Learningmanipulation}
Andrychowicz, O. A.~M., Baker, B., Chociej, M., J{\'o}zefowicz, R., McGrew, B.,
  Pachocki, J., Petron, A., Plappert, M., Powell, G., Ray, A., Schneider, J.,
  Sidor, S., Tobin, J., Welinder, P., Weng, L., and Zaremba, W. (2020).
\newblock Learning dexterous in-hand manipulation.
\newblock {\em International Journal of Robotics Research}, 39(1):3--20.

\bibitem[Argall et~al., 2011]{Argall2011TactileAdaptation}
Argall, B.~D., Sauser, E.~L., and Billard, A.~G. (2011).
\newblock Tactile guidance for policy adaptation.
\newblock {\em Foundations and Trends{\textregistered} in Robotics},
  1(2):79--133.

\bibitem[Arunachalam et~al., 2023a]{Arunachalam2023Holo-DexReality}
Arunachalam, S.~P., G{\"u}zey, I., Chintala, S., and Pinto, L. (2023a).
\newblock Holo-dex: Teaching dexterity with immersive mixed reality.
\newblock In {\em 2023 IEEE International Conference on Robotics and Automation
  (ICRA)}, volume 2023-May, pages 5962--5969, London, United Kingdom. IEEE.

\bibitem[Arunachalam et~al., 2023b]{Arunachalam2023DexterousManipulation}
Arunachalam, S.~P., Silwal, S., Evans, B., and Pinto, L. (2023b).
\newblock Dexterous imitation made easy: A learning-based framework for
  efficient dexterous manipulation.
\newblock In {\em 2023 IEEE International Conference on Robotics and Automation
  (ICRA)}, pages 5954--5961, London, United Kingdom. IEEE.

\bibitem[Ben~Amor et~al., 2012]{Amor2012Generalizationhands}
Ben~Amor, H., Kroemer, O., Hillenbrand, U., Neumann, G., and Peters, J. (2012).
\newblock Generalization of human grasping for multi-fingered robot hands.
\newblock In {\em 2012 IEEE/RSJ International Conference on Intelligent Robots
  and Systems}, pages 2043--2050, Vilamoura, Algarve, Portugal. IEEE.

\bibitem[Birglen et~al., 2008]{Birglen2008Underactuatedhands}
Birglen, L., Lalibert{\'e}, T., and Gosselin, C. (2008).
\newblock {\em Underactuated Robotic Hands}, volume~40.
\newblock Springer Berlin Heidelberg, Berlin, Heidelberg.

\bibitem[Bjorck et~al., 2025]{Bjorck2025GR00TRobots}
Bjorck, J., Casta{\~n}eda, F., Cherniadev, N., Da, X., Ding, R., Fan, L.~J.,
  Fang, Y., Fox, D., Hu, F., Huang, S., Jang, J., Jiang, Z., Kautz, J.,
  Kundalia, K., Lao, L., Li, Z., Lin, Z., Lin, K., Liu, G., Llontop, E., Magne,
  L., Mandlekar, A., Narayan, A., Nasiriany, S., Reed, S., Tan, Y.~L., Wang,
  G., Wang, Z., Wang, J., Wang, Q., Xiang, J., Xie, Y., Xu, Y., Xu, Z., Ye, S.,
  Yu, Z., Zhang, A., Zhang, H., Zhao, Y., Zheng, R., and Zhu, Y. (2025).
\newblock Gr00t n1: An open foundation model for generalist humanoid robots.

\bibitem[Bl{\"a}ttner et~al., 2023]{Bla2023DMFC-GraspNetScenes}
Bl{\"a}ttner, P., Brand, J., Neumann, G., and Vien, N.~A. (2023).
\newblock Dmfc-graspnet: Differentiable multi-fingered robotic grasp generation
  in cluttered scenes.
\newblock {\em arXiv preprint arXiv:2308.00456}.

\bibitem[{BMW AG}, 2024]{BMW2024}
{BMW AG} (2024).
\newblock Successful test of humanoid robots at bmw group plant spartanburg.
\newblock
  \url{https://www.press.bmwgroup.com/global/article/detail/T0444265EN/successful-test-of-humanoid-robots-at-bmw-group-plant-spartanburg?language=en}.
\newblock Accessed on 2024-08-22.

\bibitem[Bradley~Knox and Stone, 2008]{Knox2008TAMERReinforcement}
Bradley~Knox, W. and Stone, P. (2008).
\newblock Tamer: Training an agent manually via evaluative reinforcement.
\newblock In {\em 2008 7th IEEE International Conference on Development and
  Learning}, pages 292--297, Monterey, CA, USA. IEEE.

\bibitem[Brohan et~al., 2023]{Brohan2023RT-1Scale}
Brohan, A., Brown, N., Carbajal, J., Chebotar, Y., Dabis, J., Finn, C.,
  Gopalakrishnan, K., Hausman, K., Herzog, A., Hsu, J., Ibarz, J., Ichter, B.,
  Irpan, A., Jackson, T., Jesmonth, S., Joshi, N.~J., Julian, R., Kalashnikov,
  D., Kuang, Y., Leal, I., Lee, K.-H., Levine, S., Lu, Y., Malla, U.,
  Manjunath, D., Mordatch, I., Nachum, O., Parada, C., Peralta, J., Perez, E.,
  Pertsch, K., Quiambao, J., Rao, K., Ryoo, M., Salazar, G., Sanketi, P.,
  Sayed, K., Singh, J., Sontakke, S., Stone, A., Tan, C., Tran, H., Vanhoucke,
  V., Vega, S., Vuong, Q., Xia, F., Xiao, T., Xu, P., Xu, S., Yu, T., and
  Zitkovich, B. (2023).
\newblock Rt-1: Robotics transformer for real-world control at scale.
\newblock (arXiv:2212.06817).

\bibitem[Cai et~al., 2024]{Cai2024VisualRearrangement}
Cai, Y., Gao, J., Pohl, C., and Asfour, T. (2024).
\newblock Visual imitation learning of task-oriented object grasping and
  rearrangement.
\newblock {\em arXiv preprint arXiv:2403.14000}.

\bibitem[Celemin et~al., 2019]{Celemin2019Reinforcementadvice}
Celemin, C., Maeda, G., {Ruiz-del-Solar}, J., Peters, J., and Kober, J. (2019).
\newblock Reinforcement learning of motor skills using policy search and human
  corrective advice.
\newblock {\em The International Journal of Robotics Research},
  38(14):1560--1580.

\bibitem[Celemin et~al., 2022]{Celemin2022InteractiveSurvey}
Celemin, C., {P{\'e}rez-Dattari}, R., Chisari, E., Franzese, G., {de Souza
  Rosa}, L., Prakash, R., Ajanovi{\'c}, Z., Ferraz, M., Valada, A., and Kober,
  J. (2022).
\newblock Interactive imitation learning in robotics: A survey.
\newblock {\em Foundations and Trends{\textregistered} in Robotics},
  10(1-2):1--197.

\bibitem[Celemin and {Ruiz-del-Solar}, 2019]{Celemin2019AnFeedback}
Celemin, C. and {Ruiz-del-Solar}, J. (2019).
\newblock An interactive framework for learning continuous actions policies
  based on corrective feedback.
\newblock {\em Journal of Intelligent \& Robotic Systems}, 95(1):77--97.

\bibitem[Chen et~al., 2024]{Chen2024SpringGraspUncertainty}
Chen, S., Bohg, J., and Liu, K. (2024).
\newblock Springgrasp: Synthesizing compliant, dexterous grasps under shape
  uncertainty.
\newblock In {\em Robotics: Science and Systems XX}, Delft, Netherlands.
  {Robotics: Science and Systems Foundation}.

\bibitem[Chi et~al., 2024a]{Chi2024DiffusionDiffusion}
Chi, C., Xu, Z., Feng, S., Cousineau, E., Du, Y., Burchfiel, B., Tedrake, R.,
  and Song, S. (2024a).
\newblock Diffusion policy: Visuomotor policy learning via action diffusion.
\newblock {\em The International Journal of Robotics Research}.

\bibitem[Chi et~al., 2024b]{Chi2024UniversalRobots}
Chi, C., Xu, Z., Pan, C., Cousineau, E., Burchfiel, B., Feng, S., Tedrake, R.,
  and Song, S. (2024b).
\newblock Universal manipulation interface: In-the-wild robot teaching without
  in-the-wild robots.
\newblock {\em arXiv preprint arXiv:2402.10329}.

\bibitem[Chisari et~al., 2022]{Chisari2022CorrectManipulation}
Chisari, E., Welschehold, T., Boedecker, J., Burgard, W., and Valada, A.
  (2022).
\newblock Correct me if i am wrong: Interactive learning for robotic
  manipulation.
\newblock {\em IEEE Robotics and Automation Letters}, 7(2):3695--3702.

\bibitem[Christiano et~al., 2017]{Christiano2017DeepPreferences}
Christiano, P.~F., Leike, J., Brown, T.~B., Martic, M., Legg, S., and Amodei,
  D. (2017).
\newblock Deep reinforcement learning from human preferences.
\newblock In {I. Guyon and U. Von Luxburg and S. Bengio and H. Wallach and R.
  Fergus and S. Vishwanathan and R. Garnett}, editor, {\em Advances in Neural
  Information Processing Systems}, volume~30, Long Beach, CA, USA. Curran
  Associates, Inc.

\bibitem[Dahiya et~al., 2010]{Dahiya2010TactileHumanoids}
Dahiya, R., Metta, G., Valle, M., and Sandini, G. (2010).
\newblock Tactile sensing---from humans to humanoids.
\newblock {\em IEEE Transactions on Robotics}, 26(1):1--20.

\bibitem[{de Farias} et~al., 2024]{Farias2024Task-InformedObjects}
{de Farias}, C., Tamadazte, B., Adjigble, M., Stolkin, R., and Marturi, N.
  (2024).
\newblock Task-informed grasping of partially observed objects.
\newblock {\em IEEE Robotics and Automation Letters}, 9(10):8394--8401.

\bibitem[Diehl et~al., 2021]{Diehl2021AutomatedObservations}
Diehl, M., Paxton, C., and {Ramirez-Amaro}, K. (2021).
\newblock Automated generation of robotic planning domains from observations.
\newblock In {\em 2021 IEEE/RSJ International Conference on Intelligent Robots
  and Systems (IROS)}, pages 6732--6738, Prague, Czech Republic. IEEE.

\bibitem[Ding et~al., 2023]{Ding2023LearningPreference}
Ding, Z., Chen, Y., Ren, A.~Z., Gu, S.~S., Wang, Q., Dong, H., and Jin, C.
  (2023).
\newblock Learning a universal human prior for dexterous manipulation from
  human preference.
\newblock {\em arXiv preprint arXiv:2304.04602}.

\bibitem[Ewerton et~al., 2016]{Ewerton2016Incrementalskills}
Ewerton, M., Maeda, G., Kollegger, G., Wiemeyer, J., and Peters, J. (2016).
\newblock Incremental imitation learning of context-dependent motor skills.
\newblock In {\em 2016 IEEE-RAS 16th International Conference on Humanoid
  Robots (Humanoids)}, pages 351--358, Cancun, Mexico. IEEE.

\bibitem[Falco et~al., 2018]{Falco2018OnManipulation}
Falco, P., Attawia, A., Saveriano, M., and Lee, D. (2018).
\newblock On policy learning robust to irreversible events: An application to
  robotic in-hand manipulation.
\newblock {\em IEEE Robotics and Automation Letters}, 3(3):1482--1489.

\bibitem[Feng et~al., 2024]{Feng2024FFHFlowTime}
Feng, Q., Feng, J., Chen, Z., Triebel, R., and Knoll, A. (2024).
\newblock Ffhflow: A flow-based variational approach for multi-fingered grasp
  synthesis in real time.
\newblock {\em arXiv preprint arXiv:2407.15161}.

\bibitem[Florence et~al., 2022]{Florence2022ImplicitCloning}
Florence, P., Lynch, C., Zeng, A., Ramirez, O.~A., Wahid, A., Downs, L., Wong,
  A., Lee, J., Mordatch, I., and Tompson, J. (2022).
\newblock Implicit behavioral cloning.
\newblock In {\em Proceedings of the 5th Conference on Robot Learning}, pages
  158--168, London, UK. PMLR.

\bibitem[Freiberg et~al., 2025]{Freiberg2025DiffusionGrasping}
Freiberg, R., Qualmann, A., Vien, N.~A., and Neumann, G. (2025).
\newblock Diffusion for multi-embodiment grasping.
\newblock {\em IEEE Robotics and Automation Letters}, 10(3):2694--2701.

\bibitem[Gams et~al., 2016]{Gams2016Adaptationinteraction}
Gams, A., Petri{\v c}, T., Do, M., Nemec, B., Morimoto, J., Asfour, T., and
  Ude, A. (2016).
\newblock Adaptation and coaching of periodic motion primitives through
  physical and visual interaction.
\newblock {\em Robotics and Autonomous Systems}, 75:340--351.

\bibitem[Gao et~al., 2024]{Gao2024Bi-KVILTasks}
Gao, J., Jin, X., Krebs, F., Jaquier, N., and Asfour, T. (2024).
\newblock Bi-kvil: Keypoints-based visual imitation learning of bimanual
  manipulation tasks.
\newblock In {\em 2024 IEEE International Conference on Robotics and Automation
  (ICRA)}, pages 16850--16857, Yokohama, Japan. IEEE.

\bibitem[Gao et~al., 2023]{Gao2023K-VILLearning}
Gao, J., Tao, Z., Jaquier, N., and Asfour, T. (2023).
\newblock K-vil: Keypoints-based visual imitation learning.
\newblock {\em IEEE Transactions on Robotics}, 39(5):3888--3908.

\bibitem[Gilles et~al., 2024]{Gilles2024MetaGraspNetV2Grasping}
Gilles, M., Chen, Y., Zeng, E.~Z., Wu, Y., Furmans, K., Wong, A., and Rayyes,
  R. (2024).
\newblock Metagraspnetv2: All-in-one dataset enabling fast and reliable robotic
  bin picking via object relationship reasoning and dexterous grasping.
\newblock {\em IEEE Transactions on Automation Science and Engineering},
  21(3):2302--2320.

\bibitem[Gilles et~al., 2025]{Gilles2025MetaMVUCGrasping}
Gilles, M., Furmans, K., and Rayyes, R. (2025).
\newblock Metamvuc: Active learning for sample-efficient sim-to-real domain
  adaptation in robotic grasping.
\newblock {\em IEEE Robotics and Automation Letters}, 10(4):3644--3651.

\bibitem[Grill et~al., 2020]{Grill2020BootstrapLearning}
Grill, J.-B., Strub, F., Altch{\'e}, F., Tallec, C., Richemond, P.~H.,
  Buchatskaya, E., Doersch, C., Avila~Pires, B., Daniel~Guo, Z.,
  Gheshlaghi~Azar, M., Piot, B., Kavukcuoglu, K., Munos, R., and Valko, M.
  (2020).
\newblock Bootstrap your own latent a new approach to self-supervised learning.
\newblock In {H. Larochelle and M. Ranzato and R. Hadsell and M.F. Balcan and
  H. Lin}, editor, {\em Advances in Neural Information Processing Systems},
  volume~33, pages 21271--21284, Virtual. Curran Associates, Inc.

\bibitem[Gro{\ss} et~al., 2024]{Gross2024OPENGRASP-LITEMechanism}
Gro{\ss}, S., Ratzel, M., Welte, E., {Hidalgo-Carvajal}, D., Chen, L.,
  Fortuni{\'c}, E.~P., Ganguly, A., Swikir, A., and Haddadin, S. (2024).
\newblock Opengrasp-lite version 1.0: A tactile artificial hand with a
  compliant linkage mechanism.
\newblock In {\em 2024 IEEE/RSJ International Conference on Intelligent Robots
  and Systems (IROS)}, pages 5311--5318, Abu Dhabi, UAE. IEEE.

\bibitem[Guo et~al., 2023]{Guo2023RecentSurvey}
Guo, H., Wu, F., Qin, Y., Li, R., Li, K., and Li, K. (2023).
\newblock Recent trends in task and motion planning for robotics: A survey.
\newblock {\em ACM Computing Surveys}, 55(13s):1--36.

\bibitem[Gupta et~al., 2016]{Gupta2016Learningdemonstrations}
Gupta, A., Eppner, C., Levine, S., and Abbeel, P. (2016).
\newblock Learning dexterous manipulation for a soft robotic hand from human
  demonstrations.
\newblock In {\em 2016 IEEE/RSJ International Conference on Intelligent Robots
  and Systems (IROS)}, pages 3786--3793, Daejeon, Korea (South). IEEE.

\bibitem[Gupta et~al., 2021]{Gupta2021Reset-FreeIntervention}
Gupta, A., Yu, J., Zhao, T.~Z., Kumar, V., Rovinsky, A., Xu, K., Devlin, T.,
  and Levine, S. (2021).
\newblock Reset-free reinforcement learning via multi-task learning: Learning
  dexterous manipulation behaviors without human intervention.
\newblock In {\em 2021 IEEE International Conference on Robotics and Automation
  (ICRA)}, pages 6664--6671, Xi'an, China. IEEE.

\bibitem[Haarnoja et~al., 2018]{Haarnoja2018SoftApplications}
Haarnoja, T., Zhou, A., Hartikainen, K., Tucker, G., Ha, S., Tan, J., Kumar,
  V., Zhu, H., Gupta, A., Abbeel, P., and Levine, S. (2018).
\newblock Soft actor-critic algorithms and applications.
\newblock {\em arXiv preprint arXiv:1812.05905}.

\bibitem[Han et~al., 2023]{Han2023AManipulation}
Han, D., Mulyana, B., Stankovic, V., and Cheng, S. (2023).
\newblock A survey on deep reinforcement learning algorithms for robotic
  manipulation.
\newblock {\em Sensors}, 23(7).

\bibitem[Han et~al., 2024]{Han2024LearningObservations}
Han, Y., Chen, Z., Williams, K.~A., and Ravichandar, H. (2024).
\newblock Learning prehensile dexterity by imitating and emulating state-only
  observations.
\newblock {\em IEEE Robotics and Automation Letters}, 9(10):8266--8273.

\bibitem[Hastie et~al., 2009]{Hastie2009TheLearning}
Hastie, T., Tibshirani, R., and Friedman, J. (2009).
\newblock {\em The Elements of Statistical Learning}.
\newblock Springer New York, New York, NY.

\bibitem[He and Ciocarlie, 2022]{He2022DiscoveringLearning}
He, Z. and Ciocarlie, M. (2022).
\newblock Discovering synergies for robot manipulation with multi-task
  reinforcement learning.
\newblock In {\em 2022 International Conference on Robotics and Automation
  (ICRA)}, pages 2714--2721, Philadelphia, PA, USA. IEEE.

\bibitem[Hester et~al., 2018]{Hester2018DeepDemonstrations}
Hester, T., Vecerik, M., Pietquin, O., Lanctot, M., Schaul, T., Piot, B.,
  Horgan, D., Quan, J., Sendonaris, A., Osband, I., {Dulac-Arnold}, G.,
  Agapiou, J., Leibo, J., and Gruslys, A. (2018).
\newblock Deep q-learning from demonstrations.
\newblock {\em Proceedings of the AAAI Conference on Artificial Intelligence},
  32(1).

\bibitem[{Hidalgo-Carvajal} et~al.,
  2023]{Hidalgo-Carvajal2023AnthropomorphicCompletion}
{Hidalgo-Carvajal}, D., Chen, H., Bettelani, G.~C., Jung, J., Zavaglia, M.,
  Busse, L., Naceri, A., Leutenegger, S., and Haddadin, S. (2023).
\newblock Anthropomorphic grasping with neural object shape completion.
\newblock {\em IEEE Robotics and Automation Letters}, 8(12):8034--8041.

\bibitem[Hoque et~al., 2021]{Hoque2021LazyDAggerLearning}
Hoque, R., Balakrishna, A., Putterman, C., Luo, M., Brown, D.~S., Seita, D.,
  Thananjeyan, B., Novoseller, E., and Goldberg, K. (2021).
\newblock Lazydagger: Reducing context switching in interactive imitation
  learning.
\newblock In {\em 2021 IEEE 17th International Conference on Automation Science
  and Engineering (CASE)}, pages 502--509, Lyon, France. IEEE.

\bibitem[Hu et~al., 2022]{Hu2022LearnDemonstration}
Hu, Y., Li, K., and Wei, N. (2022).
\newblock Learn to grasp objects with dexterous robot manipulator from human
  demonstration.
\newblock In {\em 2022 International Conference on Advanced Robotics and
  Mechatronics (ICARM)}, pages 1062--1067, Guilin, China. IEEE.

\bibitem[Hu et~al., 2023]{Hu2023REBOOTManipulation}
Hu, Z., Rovinsky, A., Luo, J., Kumar, V., Gupta, A., and Levine, S. (2023).
\newblock Reboot: Reuse data for bootstrapping efficient real-world dexterous
  manipulation.
\newblock In {\em Proceedings of 7th Conference on Robot Learning}, volume 229,
  Atlanta, USA. PMLR.

\bibitem[Huang et~al., 2023]{Huang2023Dexteroushand}
Huang, L., Cai, W., Zhu, Z., and Zou, Z. (2023).
\newblock Dexterous manipulation of construction tools using anthropomorphic
  robotic hand.
\newblock {\em Automation in Construction}, 156:105133.

\bibitem[{INSPIRE-ROBOTS}, 2024]{InspireHand}
{INSPIRE-ROBOTS} (2024).
\newblock The dexterous hands rh56dfx series.
\newblock \url{https://en.inspire-robots.com/product/rh56dfx}.
\newblock Accessed on 2024-05-14.

\bibitem[Jin et~al., 2023]{Jin2023Progressinteractions}
Jin, J., Wang, S., Zhang, Z., Mei, D., and Wang, Y. (2023).
\newblock Progress on flexible tactile sensors in robotic applications on
  objects properties recognition, manipulation and human-machine interactions.
\newblock {\em Soft Science}, 3(1).

\bibitem[Kadalagere~Sampath et~al., 2023]{Sampath2023Reviewhands}
Kadalagere~Sampath, S., Wang, N., Wu, H., and Yang, C. (2023).
\newblock Review on human-like robot manipulation using dexterous hands.
\newblock {\em Cognitive Computation and Systems}, 5(1):14--29.

\bibitem[Kahn et~al., 2021]{Kahn2021LaNDDisengagements}
Kahn, G., Abbeel, P., and Levine, S. (2021).
\newblock Land: Learning to navigate from disengagements.
\newblock {\em IEEE Robotics and Automation Letters}, 6(2):1872--1879.

\bibitem[Kaya and Oztop, 2018]{Kaya2018EffectiveControl}
Kaya, O. and Oztop, E. (2018).
\newblock Effective robot skill synthesis via divided control.
\newblock In {\em 2018 IEEE International Conference on Robotics and
  Biomimetics (ROBIO)}, pages 766--771, Kuala Lumpur, Malaysia. IEEE.

\bibitem[Kelly et~al., 2019]{Kelly2019HG-DAggerExperts}
Kelly, M., Sidrane, C., {Driggs-Campbell}, K., and Kochenderfer, M.~J. (2019).
\newblock Hg-dagger: Interactive imitation learning with human experts.
\newblock In {\em 2019 International Conference on Robotics and Automation
  (ICRA)}, pages 8077--8083, Montreal, QC, Canada. IEEE.

\bibitem[Kubus et~al., 2018]{Kubus2018LearningEfficiently}
Kubus, D., Rayyes, R., and Steil, J.~J. (2018).
\newblock Learning forward and inverse kinematics maps efficiently.
\newblock In {\em 2018 IEEE/RSJ International Conference on Intelligent Robots
  and Systems (IROS)}, pages 5133--5140, Madrid, Spain. IEEE.

\bibitem[Kumar et~al., 2019]{Kumar2019ContextualPolicies}
Kumar, V., Hermans, T., Fox, D., Birchfield, S., and Tremblay, J. (2019).
\newblock Contextual reinforcement learning of visuo-tactile multi-fingered
  grasping policies.
\newblock {\em arXiv preprint arXiv:1911.09233}.

\bibitem[Li et~al., 2024]{Li2024ContinualManipulation}
Li, L., Donato, E., Lomonaco, V., and Falotico, E. (2024).
\newblock Continual policy distillation of reinforcement learning-based
  controllers for soft robotic in-hand manipulation.
\newblock In {\em 2024 IEEE 7th International Conference on Soft Robotics
  (RoboSoft)}, pages 1026--1033, San Diego, CA, USA. IEEE.

\bibitem[Li et~al., 2023]{Li2023Within-HandLearning}
Li, X., Chen, W., Wang, Y., Diao, Q., Wu, S., and Yang, F. (2023).
\newblock Within-hand manipulation with an underactuated dexterous hand based
  on pre-trained reinforcement learning.
\newblock In {\em 2023 China Automation Congress (CAC)}, pages 3171--3176,
  Chongqing, China. IEEE.

\bibitem[Liconti et~al., 2024]{Liconti2024LeveragingHand}
Liconti, D., Toshimitsu, Y., and Katzschmann, R. (2024).
\newblock Leveraging pretrained latent representations for few-shot imitation
  learning on an anthropomorphic robotic hand.
\newblock In {\em 2024 IEEE-RAS 23rd International Conference on Humanoid
  Robots (Humanoids)}, pages 181--188, Nancy, France. IEEE.

\bibitem[Liu et~al., 2025a]{Liu2025RoboDexVLMManipulation}
Liu, H., Guo, S., Mai, P., Cao, J., Li, H., and Ma, J. (2025a).
\newblock Robodexvlm: Visual language model-enabled task planning and motion
  control for dexterous robot manipulation.
\newblock (arXiv:2503.01616).

\bibitem[Liu et~al., 2024]{Liu2024Robotdeployment}
Liu, H., Nasiriany, S., Zhang, L., Bao, Z., and Zhu, Y. (2024).
\newblock Robot learning on the job: Human-in-the-loop autonomy and learning
  during deployment.
\newblock {\em The International Journal of Robotics Research}, page
  02783649241273901.

\bibitem[Liu et~al., 2025b]{Liu2025VTDexManiplearning}
Liu, Q., Cui, Y., Sun, Z., Li, G., Chen, J., and Ye, Q. (2025b).
\newblock Vtdexmanip: a dataset and benchmark for visual-tactile pretraining
  and dexterous manipulation with reinforcement learning.
\newblock In Yue, Y., Garg, A., Peng, N., Sha, F., and Yu, R., editors, {\em
  International conference on representation learning}, volume 2025, pages
  90582--90607.

\bibitem[Lu et~al., 2024]{Lu2024OvercomingFactorization}
Lu, C., Shi, L., Chen, Z., Wu, C., and Wierman, A. (2024).
\newblock Overcoming the curse of dimensionality in reinforcement learning
  through approximate factorization.
\newblock {\em arXiv preprint arXiv:2411.07591}.

\bibitem[Luo et~al., 2023]{Luo2023RLIFLearning}
Luo, J., Dong, P., Zhai, Y., Ma, Y., and Levine, S. (2023).
\newblock Rlif: Interactive imitation learning as reinforcement learning.
\newblock {\em arXiv preprint arXiv:2311.12996}.

\bibitem[Luo et~al., 2021]{Luo2021RobustStudy}
Luo, J., Sushkov, O., Pevceviciute, R., Lian, W., Su, C., Vecerik, M., Ye, N.,
  Schaal, S., and Scholz, J. (2021).
\newblock Robust multi-modal policies for industrial assembly via reinforcement
  learning and demonstrations: A large-scale study.
\newblock In {\em Robotics: Science and Systems XVII}, Virtual. {Robotics:
  Science and Systems Foundation}.

\bibitem[Macglashan et~al., 2017]{Macglashan2017InteractiveFeedback}
Macglashan, J., Ho, M.~K., Loftin, R., Peng, B., Wang, G., Roberts, D.~L.,
  Taylor, M.~E., and Littman, M.~L. (2017).
\newblock Interactive learning from policy-dependent human feedback.
\newblock In {\em Proceedings of the 34th International Conference on Machine
  Learning}, pages 2285--2294, Sydney, Australia. PMLR.

\bibitem[Mandikal and Grauman, 2022]{Mandikal2022DexVIPVideo}
Mandikal, P. and Grauman, K. (2022).
\newblock Dexvip: Learning dexterous grasping with human hand pose priors from
  video.
\newblock In Faust, A., Hsu, D., and Neumann, G., editors, {\em Proceedings of
  the 5th Conference on Robot Learning}, pages 651--661, London, UK. PMLR.

\bibitem[Mandlekar et~al., 2020]{Mandlekar2020Human-in-the-LoopTeleoperation}
Mandlekar, A., Xu, D., {Mart{\'i}n-Mart{\'i}n}, R., Zhu, Y., {Fei-Fei}, L., and
  Savarese, S. (2020).
\newblock Human-in-the-loop imitation learning using remote teleoperation.
\newblock {\em arXiv preprint arXiv:2012.06733}.

\bibitem[Mao et~al., 2024]{Mao2024DexSkillsTasks}
Mao, X., Giudici, G., Coppola, C., Althoefer, K., Farkhatdinov, I., Li, Z., and
  Jamone, L. (2024).
\newblock Dexskills: Skill segmentation using haptic data for learning
  autonomous long-horizon robotic manipulation tasks.
\newblock In {\em 2024 IEEE/RSJ International Conference on Intelligent Robots
  and Systems (IROS)}, pages 5104--5111, Abu Dhabi, United Arab Emirates. IEEE.

\bibitem[Mees et~al., 2022]{Mees2022WhatData}
Mees, O., Hermann, L., and Burgard, W. (2022).
\newblock What matters in language conditioned robotic imitation learning over
  unstructured data.
\newblock {\em IEEE Robotics and Automation Letters}, 7(4):11205--11212.

\bibitem[Melchiorri and Kaneko, 2016]{Melchiorri2016RobotHands}
Melchiorri, C. and Kaneko, M. (2016).
\newblock Robot hands.
\newblock In {\em Springer Handbook of Robotics}, pages 463--480. Springer
  International Publishing, Cham.

\bibitem[Melnik et~al., 2019]{Melnik2019TactileTasks}
Melnik, A., Lach, L., Plappert, M., Korthals, T., Haschke, R., and Ritter, H.
  (2019).
\newblock Tactile sensing and deep reinforcement learning for in-hand
  manipulation tasks.
\newblock In {\em IROS Workshop on Autonomous Object Manipulation}, volume~39,
  pages 3--20, Macau, China. IEEE.

\bibitem[Mikami, 2009]{Mikami2009ImitationLearning}
Mikami, A. (2009).
\newblock Imitation learning.
\newblock In {\em Encyclopedia of Neuroscience}, pages 1915--1918. Springer
  Berlin Heidelberg, Berlin, Heidelberg.

\bibitem[{mimic robotics AG}, 2024]{mimic}
{mimic robotics AG} (2024).
\newblock mimic.
\newblock \url{https://www.mimicrobotics.com/}.
\newblock Accessed on 2024-08-22.

\bibitem[Mosbach et~al., 2022]{Mosbach2022AcceleratingDemonstrations}
Mosbach, M., Moraw, K., and Behnke, S. (2022).
\newblock Accelerating interactive human-like manipulation learning with
  gpu-based simulation and high-quality demonstrations.
\newblock In {\em 2022 IEEE-RAS 21st International Conference on Humanoid
  Robots (Humanoids)}, volume 2022-November, pages 435--441, Ginowan, Japan.
  IEEE.

\bibitem[Nair et~al., 2020]{Nair2020AWACDatasets}
Nair, A., Gupta, A., Dalal, M., and Levine, S. (2020).
\newblock Awac: Accelerating online reinforcement learning with offline
  datasets.
\newblock {\em arXiv preprint arXiv:2006.09359}.

\bibitem[Nair et~al., 2018]{Nair2018OvercomingDemonstrations}
Nair, A., McGrew, B., Andrychowicz, M., Zaremba, W., and Abbeel, P. (2018).
\newblock Overcoming exploration in reinforcement learning with demonstrations.
\newblock In {\em 2018 IEEE International Conference on Robotics and Automation
  (ICRA)}, pages 6292--6299, Brisbane, QLD, Australia. IEEE.

\bibitem[Nair et~al., 2022]{Nair2022R3MManipulation}
Nair, S., Rajeswaran, A., Kumar, V., Finn, C., and Gupta, A. (2022).
\newblock R3m: A universal visual representation for robot manipulation.
\newblock (arXiv:2203.12601).

\bibitem[Najar and Chetouani, 2021]{Najar2021ReinforcementSurvey}
Najar, A. and Chetouani, M. (2021).
\newblock Reinforcement learning with human advice: A survey.
\newblock {\em Frontiers in Robotics and AI}, 8.

\bibitem[Okamura et~al., 2000]{Okamura2000Anmanipulation}
Okamura, A., Smaby, N., and Cutkosky, M. (2000).
\newblock An overview of dexterous manipulation.
\newblock In {\em Proceedings 2000 ICRA. Millennium Conference. IEEE
  International Conference on Robotics and Automation. Symposia Proceedings
  (Cat. No.00CH37065)}, volume~1, pages 255--262, San Francisco, CA, USA. IEEE.

\bibitem[Orbik et~al., 2021]{Orbik2021Inversemanipulation}
Orbik, J., Agostini, A., and Lee, D. (2021).
\newblock Inverse reinforcement learning for dexterous hand manipulation.
\newblock In {\em 2021 IEEE International Conference on Development and
  Learning (ICDL)}, pages 1--7, Beijing, China. IEEE.

\bibitem[Osa et~al., 2018]{Osa2018AnLearning}
Osa, T., Pajarinen, J., Neumann, G., Bagnell, J.~A., Abbeel, P., and Peters, J.
  (2018).
\newblock An algorithmic perspective on imitation learning.
\newblock {\em Foundations and Trends in Robotics}, 7(1--2):1--179.

\bibitem[Parnichkun et~al., 2022]{Parnichkun2022ReILLearning}
Parnichkun, R., Dailey, M.~N., and Yamashita, A. (2022).
\newblock Reil: A framework for reinforced intervention-based imitation
  learning.
\newblock {\em arXiv preprint arXiv:2203.15390}.

\bibitem[{PaXini}, 2025]{PaXiniDexH13GEN2}
{PaXini} (2025).
\newblock Paxini dexh13gen2.
\newblock \url{https://paxini.com/}.
\newblock Accessed on 2025-07-25.

\bibitem[Pearce et~al., 2023]{Pearce2023ImitatingModels}
Pearce, T., Rashid, T., Kanervisto, A., Bignell, D., Sun, M., Georgescu, R.,
  Macua, S.~V., Tan, S.~Z., Momennejad, I., Hofmann, K., and Devlin, S. (2023).
\newblock Imitating human behaviour with diffusion models.
\newblock {\em arXiv preprint arXiv:2301.10677}.

\bibitem[{P{\'e}rez-Dattari} et~al., 2020]{Pe2020InteractiveNetworks}
{P{\'e}rez-Dattari}, R., Celemin, C., {Ruiz-del-Solar}, J., and Kober, J.
  (2020).
\newblock Interactive learning with corrective feedback for policies based on
  deep neural networks.
\newblock In {\em Proceedings of the 2018 International Symposium on
  Experimental Robotics}, pages 353--363. Springer International Publishing,
  Cham.

\bibitem[{PRENSILIA}, 2023]{PrensiliaHand}
{PRENSILIA} (2023).
\newblock Ih2 azzurra - prensilia - grasping innovation.
\newblock \url{https://www.prensilia.com/ih2-azzurra-hand/}.
\newblock Accessed on 2024-05-14.

\bibitem[{qbrobotics}, 2022]{SoftHand2}
{qbrobotics} (2022).
\newblock qb softhand2 research - qbrobotics.
\newblock \url{https://qbrobotics.com/product/qb-softhand-2-research/}.
\newblock Accessed on 2024-05-14.

\bibitem[Qin et~al., 2022a]{Qin2022FromTeleoperation}
Qin, Y., Su, H., and Wang, X. (2022a).
\newblock From one hand to multiple hands: Imitation learning for dexterous
  manipulation from single-camera teleoperation.
\newblock {\em IEEE Robotics and Automation Letters}, 7(4):10873--10881.

\bibitem[Qin et~al., 2022b]{Qin2022DexMVVideos}
Qin, Y., Wu, Y.-H., Liu, S., Jiang, H., Yang, R., Fu, Y., and Wang, X. (2022b).
\newblock Dexmv: Imitation learning for~dexterous manipulation from~human
  videos.
\newblock {\em Lecture Notes in Computer Science (including subseries Lecture
  Notes in Artificial Intelligence and Lecture Notes in Bioinformatics)}, 13699
  LNCS:570--587.

\bibitem[Radosavovic et~al., 2021]{Radosavovic2021State-OnlyManipulation}
Radosavovic, I., Wang, X., Pinto, L., and Malik, J. (2021).
\newblock State-only imitation learning for dexterous manipulation.
\newblock In {\em 2021 IEEE/RSJ International Conference on Intelligent Robots
  and Systems (IROS)}, pages 7865--7871, Prague, Czech Republic. IEEE.

\bibitem[Rajeswaran et~al., 2018]{Rajeswaran2018LearningDemonstrations}
Rajeswaran, A., Kumar, V., Gupta, A., Vezzani, G., Schulman, J., Todorov, E.,
  and Levine, S. (2018).
\newblock Learning complex dexterous manipulation with deep reinforcement
  learning and demonstrations.
\newblock In {\em Robotics: Science and Systems XIV}, Pittsburgh, Pennsylvania.
  {Robotics: Science and Systems Foundation}.

\bibitem[{Ramirez-Amaro} et~al., 2017]{Ramirez-Amaro2017Transferringactivities}
{Ramirez-Amaro}, K., Beetz, M., and Cheng, G. (2017).
\newblock Transferring skills to humanoid robots by extracting semantic
  representations from observations of human activities.
\newblock {\em Artificial Intelligence}, 247:95--118.

\bibitem[Rayyes, 2021]{Rayyes2021EfficientRobots}
Rayyes, R. (2021).
\newblock {\em Efficient and Stable Online Learning for Developmental Robots}.
\newblock PhD thesis, TU Braunschweig.

\bibitem[Rayyes et~al., 2023]{Rayyes2023Interest-DrivenRobots}
Rayyes, R., Donat, H., Steil, J., and Spranger, M. (2023).
\newblock Interest-driven exploration with observational learning for
  developmental robots.
\newblock {\em IEEE Transactions on Cognitive and Developmental Systems},
  15(2):373--384.

\bibitem[Reuss et~al., 2024]{Reuss2024MultimodalGoals}
Reuss, M., Ya{\u g}murlu, {\"O}., Wenzel, F., and Lioutikov, R. (2024).
\newblock Multimodal diffusion transformer: Learning versatile behavior from
  multimodal goals.
\newblock In {\em Robotics: Science and Systems XX}, Delft, Netherlands.
  {Robotics: Science and Systems Foundation}.

\bibitem[{Reuters}, 2024]{Reuters2024}
{Reuters} (2024).
\newblock Tesla to have humanoid robots for internal use next year, musk says.
\newblock
  \url{https://www.reuters.com/business/autos-transportation/tesla-have-humanoid-robots-internal-use-next-year-musk-says-2024-07-22/}.
\newblock Accessed on 2024-08-22.

\bibitem[{ROBOTERA}, 2025]{XHAND1}
{ROBOTERA} (2025).
\newblock Xhand1.
\newblock \url{https://www.robotera.com/en/goods1/4.html}.
\newblock Accessed on 2025-07-03.

\bibitem[Ross et~al., 2011]{Ross2011ALearning}
Ross, S., Gordon, G., and Bagnell, D. (2011).
\newblock A reduction of imitation learning and structured prediction to
  no-regret online learning.
\newblock {\em Proceedings of the Fourteenth International Conference on
  Artificial Intelligence and Statistics}, 15:627--635.

\bibitem[Ruppel and Zhang, 2020]{Ruppel2020LearningDemonstration}
Ruppel, P. and Zhang, J. (2020).
\newblock Learning object manipulation with dexterous hand-arm systems from
  human demonstration.
\newblock In {\em 2020 IEEE/RSJ International Conference on Intelligent Robots
  and Systems (IROS)}, pages 5417--5424, Las Vegas, NV, USA. IEEE.

\bibitem[Russell and Norvig, 2016]{Russell2016Artificialapproach}
Russell, S.~J. and Norvig, P. (2016).
\newblock {\em Artificial intelligence: a modern approach}.
\newblock Pearson, Boston.

\bibitem[Santello et~al., 2016]{Santello2016Handhands}
Santello, M., Bianchi, M., Gabiccini, M., Ricciardi, E., Salvietti, G.,
  Prattichizzo, D., Ernst, M., Moscatelli, A., J{\"o}rntell, H., Kappers,
  A.~M., Kyriakopoulos, K., {Albu-Sch{\"a}ffer}, A., Castellini, C., and
  Bicchi, A. (2016).
\newblock Hand synergies: Integration of robotics and neuroscience for
  understanding the control of biological and artificial hands.
\newblock {\em Physics of Life Reviews}, 17:1--23.

\bibitem[{Sarcomere Dynamics}, 2024]{Artus}
{Sarcomere Dynamics} (2024).
\newblock Artus lite.
\newblock \url{https://sarcomeredynamics.com/products}.
\newblock Accessed on 2024-08-22.

\bibitem[Sauser et~al., 2012]{Sauser2012Iterativecorrections}
Sauser, E.~L., Argall, B.~D., Metta, G., and Billard, A.~G. (2012).
\newblock Iterative learning of grasp adaptation through human corrections.
\newblock {\em Robotics and Autonomous Systems}, 60(1):55--71.

\bibitem[Saveriano et~al., 2023]{Saveriano2023DynamicSurvey}
Saveriano, M., {Abu-Dakka}, F.~J., Kramberger, A., and Peternel, L. (2023).
\newblock Dynamic movement primitives in robotics: A tutorial survey.
\newblock {\em The International Journal of Robotics Research},
  42(13):1133--1184.

\bibitem[Savescu et~al., 2004]{Savescu2004ASimulation}
Savescu, A.-V., Cheze, L., Wang, X., Beurier, G., and Verriest, J.-P. (2004).
\newblock A 25 degrees of freedom hand geometrical model for better hand
  attitude simulation.
\newblock {\em Digital Human Modeling for Design and Engineering Symposium}.

\bibitem[{SCHUNK}, 2023]{SchunkSVH}
{SCHUNK} (2023).
\newblock Svh 5-finger servo-electric gripping hand.
\newblock
  \url{https://schunk.com/de/en/gripping-systems/special-gripper/svh/c/PGR\_3161}.
\newblock Accessed on 2024-05-14.

\bibitem[{seed robotics}, 2021]{SeedHand}
{seed robotics} (2021).
\newblock Rh8d adult size dexterous robot hand --- seed robotics.
\newblock \url{https://www.seedrobotics.com/rh8d-adult-robot-hand}.
\newblock Accessed on 2024-05-14.

\bibitem[{Shadow Robot Company}, 2024]{ShadowHand}
{Shadow Robot Company} (2024).
\newblock Shadow dexterous hand series - research and development tool.
\newblock \url{https://www.shadowrobot.com/dexterous-hand-series/}.
\newblock Accessed on 2024-05-14.

\bibitem[Shafiullah et~al., 2022]{Shafiullah2022Behaviorstone}
Shafiullah, N.~M., Cui, Z., Altanzaya, A.~A., and Pinto, L. (2022).
\newblock Behavior transformers: Cloning \$k\$ modes with one stone.
\newblock {\em Advances in Neural Information Processing Systems},
  35:22955--22968.

\bibitem[Shaw et~al., 2023a]{Shaw2023LEAPLearning}
Shaw, K., Agarwal, A., and Pathak, D. (2023a).
\newblock Leap hand: Low-cost, efficient, and anthropomorphic hand for robot
  learning.
\newblock {\em arXiv preprint arXiv:2309.06440}.

\bibitem[Shaw et~al., 2023b]{Shaw2023VideoDexVideos}
Shaw, K., Bahl, S., and Pathak, D. (2023b).
\newblock Videodex: Learning dexterity from internet videos.
\newblock {\em Proceedings of Machine Learning Research}, 205:654--665.

\bibitem[Shaw et~al., 2024]{Shaw2024Learningvideos}
Shaw, K., Bahl, S., Sivakumar, A., Kannan, A., and Pathak, D. (2024).
\newblock Learning dexterity from human hand motion in internet videos.
\newblock {\em The International Journal of Robotics Research}, 43(4):513--532.

\bibitem[Si et~al., 2024]{Si2024TildeDeltaHand}
Si, Z., Zhang, K.~L., Temel, Z., and Kroemer, O. (2024).
\newblock Tilde: Teleoperation for dexterous in-hand manipulation learning with
  a deltahand.
\newblock {\em arXiv preprint arXiv:2405.18804}.

\bibitem[Sivakumar et~al., 2022]{Sivakumar2022RoboticYoutube}
Sivakumar, A., Shaw, K., and Pathak, D. (2022).
\newblock Robotic telekinesis: Learning a robotic hand imitator by watching
  humans on youtube.

\bibitem[Spencer et~al., 2020]{Spencer2020Learningfeedback}
Spencer, J., Choudhury, S., Barnes, M., Schmittle, M., Chiang, M., Ramadge, P.,
  and Srinivasa, S. (2020).
\newblock Learning from interventions: Human-robot interaction as both explicit
  and implicit feedback.
\newblock In {\em Robotics: Science and Systems XVI}, Corvalis, Oregon, USA.
  {Robotics: Science and Systems Foundation}.

\bibitem[Starke and Asfour, 2024]{Starke2024KinematicGeneration}
Starke, J. and Asfour, T. (2024).
\newblock Kinematic synergy primitives for human-like grasp motion generation.
\newblock In {\em 2024 IEEE International Conference on Robotics and Automation
  (ICRA)}, pages 4119--4125, Yokohama, Japan. IEEE.

\bibitem[Sun et~al., 2023]{Sun2023MEGA-DAggerExperts}
Sun, X., Yang, S., and Mangharam, R. (2023).
\newblock Mega-dagger: Imitation learning with multiple imperfect experts.

\bibitem[Sutton and Barto, 2018]{Sutton2018ReinforcementIntroduction}
Sutton, R.~S. and Barto, A.~G. (2018).
\newblock {\em Reinforcement Learning: An Introduction}.
\newblock A Bradford Book, Cambridge, MA, USA.

\bibitem[{TESOLLO}, 2025]{tesolloDG5FHumanoidRobotic2025}
{TESOLLO} (2025).
\newblock Dg-5f {\textbar} humanoid robotic hand for dexterous manipulation.
\newblock Accessed on 2025-08-07.

\bibitem[Todorov et~al., 2012]{Todorov2012MuJoCocontrol}
Todorov, E., Erez, T., and Tassa, Y. (2012).
\newblock Mujoco: A physics engine for model-based control.
\newblock In {\em 2012 IEEE/RSJ International Conference on Intelligent Robots
  and Systems}, pages 5026--5033, Vilamoura, Algarve, Portugal. IEEE.

\bibitem[Toshimitsu et~al., 2023]{Toshimitsu2023GettingJoints}
Toshimitsu, Y., Forrai, B., Cangan, B.~G., Steger, U., Knecht, M., Weirich, S.,
  and Katzschmann, R.~K. (2023).
\newblock Getting the ball rolling: Learning a dexterous policy for a
  biomimetic tendon-driven hand with rolling contact joints.
\newblock In {\em 2023 IEEE-RAS 22nd International Conference on Humanoid
  Robots (Humanoids)}, pages 1--7, Austin, TX, USA. IEEE.

\bibitem[Ugur et~al., 2011]{Ugur2011Learningscaffolding}
Ugur, E., Celikkanat, H., Sahin, E., Nagai, Y., and Oztop, E. (2011).
\newblock Learning to grasp with parental scaffolding.
\newblock In {\em 2011 11th IEEE-RAS International Conference on Humanoid
  Robots}, pages 480--486, Bled, Slovenia. IEEE.

\bibitem[Unitree, 2025]{unitreeUnitreeDex51Smart2025}
Unitree (2025).
\newblock Unitree dex5-1 smart adaptability, instant responsiveness - unitree
  robotics.
\newblock \url{https://www.unitree.com/Dex5-1}.
\newblock Accessed on 2025-08-07.

\bibitem[Urain et~al., 2024]{Urain2024DeepDemonstrations}
Urain, J., Mandlekar, A., Du, Y., Shafiullah, M., Xu, D., Fragkiadaki, K.,
  Chalvatzaki, G., and Peters, J. (2024).
\newblock Deep generative models in robotics: A survey on learning from
  multimodal demonstrations.
\newblock (arXiv:2408.04380).

\bibitem[{van Hoof} et~al., 2015]{Hoof2015Learningfeatures}
{van Hoof}, H., Hermans, T., Neumann, G., and Peters, J. (2015).
\newblock Learning robot in-hand manipulation with tactile features.
\newblock In {\em 2015 IEEE-RAS 15th International Conference on Humanoid
  Robots (Humanoids)}, pages 121--127, Seoul, Korea (South). IEEE.

\bibitem[Wakabayashi et~al., 2024]{Wakabayashi2024Behavioralrate}
Wakabayashi, S., Kawaharazuka, K., Okada, K., and Inaba, M. (2024).
\newblock Behavioral learning of dish rinsing and scrubbing based on
  interruptive direct teaching considering assistance rate.
\newblock {\em Advanced Robotics}, 38(15):1052--1065.

\bibitem[Wang et~al., 2024a]{Wang2024DexCapManipulation}
Wang, C., Shi, H., Wang, W., Zhang, R., {Fei-Fei}, L., and Liu, K. (2024a).
\newblock Dexcap: Scalable and portable mocap data collection system for
  dexterous manipulation.
\newblock In {\em Robotics: Science and Systems XX}, Delft, Netherlands.
  {Robotics: Science and Systems Foundation}.

\bibitem[Wang et~al., 2024b]{Wang2024Multi-StageManipulation}
Wang, D., Liu, C., Chang, F., Huan, H., and Cheng, K. (2024b).
\newblock Multi-stage reinforcement learning for non-prehensile manipulation.
\newblock {\em IEEE Robotics and Automation Letters}, 9(7):6712--6719.

\bibitem[Wang et~al., 2022]{Wang2022AnTasks}
Wang, Y., {Beltran-Hernandez}, C.~C., Wan, W., and Harada, K. (2022).
\newblock An adaptive imitation learning framework for robotic complex
  contact-rich insertion tasks.
\newblock {\em Frontiers in Robotics and AI}, 8.

\bibitem[Wolf et~al., 2025]{Wolf2025DiffusionSurvey}
Wolf, R., Shi, Y., Liu, S., and Rayyes, R. (2025).
\newblock Diffusion models for robotic manipulation: A survey.
\newblock (arXiv:2504.08438).

\bibitem[{WONIK ROBOTICS}, 2023]{AllegroHand}
{WONIK ROBOTICS} (2023).
\newblock Allegro hand v4.0 - allegro hand.
\newblock
  \url{http://wiki.wonikrobotics.com/AllegroHandWiki/index.php/Allegro\_Hand\_v4.0}.
\newblock Accessed on 2024-05-14.

\bibitem[Wu et~al., 2025]{Wu2025RoboCopilotManipulation}
Wu, P., Shentu, Y., Liao, Q., Jin, D., Guo, M., Sreenath, K., Lin, X., and
  Abbeel, P. (2025).
\newblock Robocopilot: Human-in-the-loop interactive imitation learning for
  robot manipulation.

\bibitem[Yi et~al., 2022]{Yi2022AnthropomorphicLearning}
Yi, J.-B., Kim, J., Kang, T., Song, D., Park, J., and Yi, S.-J. (2022).
\newblock Anthropomorphic grasping of complex-shaped objects using imitation
  learning.
\newblock {\em Applied Sciences}, 12(24).

\bibitem[Yu and Wang, 2022]{Yu2022DexterousReview}
Yu, C. and Wang, P. (2022).
\newblock Dexterous manipulation for multi-fingered robotic hands with
  reinforcement learning: A review.
\newblock {\em Frontiers in Neurorobotics}, 16:861825.

\bibitem[Zarzoura et~al., 2019]{Zarzoura2019Investigationfunctions}
Zarzoura, M., {del Moral}, P., Awad, M.~I., and Tolbah, F.~A. (2019).
\newblock Investigation into reducing anthropomorphic hand degrees of freedom
  while maintaining human hand grasping functions.
\newblock {\em Proceedings of the Institution of Mechanical Engineers, Part H:
  Journal of Engineering in Medicine}, 233(2):279--292.

\bibitem[Ze et~al., 2023]{Ze2023H-InDexManipulation}
Ze, Y., Liu, Y., Shi, R., Qin, J., Yuan, Z., Wang, J., and Xu, H. (2023).
\newblock H-index: Visual reinforcement learning with hand-informed
  representations for dexterous manipulation.
\newblock In Oh, A., Naumann, T., Globerson, A., Saenko, K., Hardt, M., and
  Levine, S., editors, {\em Advances in Neural Information Processing Systems},
  volume~36, pages 74394--74409, New Orleans, Louisiana, USA. Curran
  Associates, Inc.

\bibitem[Ze et~al., 2024]{Ze20243DRepresentations}
Ze, Y., Zhang, G., Zhang, K., Hu, C., Wang, M., and Xu, H. (2024).
\newblock 3d diffusion policy: Generalizable visuomotor policy learning via
  simple 3d representations.
\newblock In {\em Robotics: Science and Systems XX}, Delft, Netherlands.
  {Robotics: Science and Systems Foundation}.

\bibitem[Zhang et~al., 2025]{Zhang2025FlowPolicyManipulation}
Zhang, Q., Liu, Z., Fan, H., Liu, G., Zeng, B., and Liu, S. (2025).
\newblock Flowpolicy: Enabling fast and robust 3d flow-based policy via
  consistency flow matching for robot manipulation.
\newblock {\em Proceedings of the AAAI Conference on Artificial Intelligence},
  39(14):14754--14762.

\bibitem[Zhao et~al., 2024]{Zhao2024GrainGraspGuidance}
Zhao, F., Tsetserukou, D., and Liu, Q. (2024).
\newblock Graingrasp: Dexterous grasp generation with fine-grained contact
  guidance.
\newblock In {\em 2024 IEEE International Conference on Robotics and Automation
  (ICRA)}, pages 6470--6476, Yokohama, Japan. IEEE.

\bibitem[Zhao et~al., 2023]{Zhao2023LearningHardware}
Zhao, T.~Z., Kumar, V., Levine, S., and Finn, C. (2023).
\newblock Learning fine-grained bimanual manipulation with low-cost hardware.
\newblock (arXiv:2304.13705).

\bibitem[Zhong et~al., 2025]{Zhong2025DexGraspVLAGrasping}
Zhong, Y., Huang, X., Li, R., Zhang, C., Liang, Y., Yang, Y., and Chen, Y.
  (2025).
\newblock Dexgraspvla: A vision-language-action framework towards general
  dexterous grasping.
\newblock (arXiv:2502.20900).

\bibitem[Zhu et~al., 2019]{Zhu2019DexterousLow-Cost}
Zhu, H., Gupta, A., Rajeswaran, A., Levine, S., and Kumar, V. (2019).
\newblock Dexterous manipulation with deep reinforcement learning: Efficient,
  general, and low-cost.
\newblock In {\em 2019 International Conference on Robotics and Automation
  (ICRA)}, pages 3651--3657, Montreal, QC, Canada. IEEE.

\bibitem[Zhu et~al., 2023]{Zhu2023VIOLAPriors}
Zhu, Y., Joshi, A., Stone, P., and Zhu, Y. (2023).
\newblock Viola: Imitation learning for vision-based manipulation with object
  proposal priors.
\newblock In {\em Proceedings of The 6th Conference on Robot Learning}, pages
  1199--1210, Auckland, NZ. PMLR.

\bibitem[Zitkovich et~al., 2023]{Zitkovich2023RT-2Control}
Zitkovich, B., Yu, T., Xu, S., Xu, P., Xiao, T., Xia, F., Wu, J., Wohlhart, P.,
  Welker, S., Wahid, A., Vuong, Q., Vanhoucke, V., Tran, H., Soricut, R.,
  Singh, A., Singh, J., Sermanet, P., Sanketi, P.~R., Salazar, G., Ryoo, M.~S.,
  Reymann, K., Rao, K., Pertsch, K., Mordatch, I., Michalewski, H., Lu, Y.,
  Levine, S., Lee, L., Lee, T.-W.~E., Leal, I., Kuang, Y., Kalashnikov, D.,
  Julian, R., Joshi, N.~J., Irpan, A., Ichter, B., Hsu, J., Herzog, A.,
  Hausman, K., Gopalakrishnan, K., Fu, C., Florence, P., Finn, C., Dubey,
  K.~A., Driess, D., Ding, T., Choromanski, K.~M., Chen, X., Chebotar, Y.,
  Carbajal, J., Brown, N., Brohan, A., Arenas, M.~G., and Han, K. (2023).
\newblock Rt-2: Vision-language-action models transfer web knowledge to robotic
  control.
\newblock In {\em Proceedings of The 7th Conference on Robot Learning}, pages
  2165--2183, Atlanta, USA. PMLR.

\end{thebibliography}

\end{document}